\documentclass[letterpaper, 10 pt, conference]{ieeeconf}

\IEEEoverridecommandlockouts 
\usepackage{graphics} 
\usepackage{epsfig} 
\usepackage{mathptmx} 
\usepackage{amsmath} 
\usepackage{amssymb}  
\usepackage{color,soul}
\usepackage{multicol}
\usepackage{balance}
\usepackage{verbatim}
\usepackage[bookmarks=true]{hyperref}
\usepackage{comment}
\usepackage{float}
\usepackage{multirow}
\usepackage{multicol}

\usepackage{siunitx}
\DeclareSIUnit{\nothing}{\relax}

\usepackage{gensymb}

\usepackage{bm}
\usepackage{mathtools}

\usepackage{booktabs, makecell, tabularx}

\usepackage{todonotes}

\usepackage[font=footnotesize]{subcaption}
\usepackage{flushend}
\usepackage{balance}

\usepackage[table]{xcolor}
\usepackage{tabularx} 
\usepackage{ninecolors}

\makeatletter
\let\c@rownum\relax
\let\therownum\relax
\makeatother
\usepackage{tabularray}

\usepackage{tikz}
\newcommand*\circled[1]{\tikz[baseline=(char.base)]{
		\node[shape=circle,draw,inner sep=0.2pt] (char) {#1};}}

\makeatletter
\let\NAT@parse\undefined
\makeatother
\usepackage[numbers,sort&compress]{natbib}

\newcommand{\circlednum}[1]{\textcircled{\raisebox{-0.9pt}{#1}}}

\newcommand{\videoone}{\href{https://drive.google.com/file/d/1wCQXObbLN4_GTkawG8lpD_AP3l8szFth/view?usp=drive_link}{Extension 1}}
\newcommand{\videotwo}{\href{https://drive.google.com/file/d/13yr67N18D8QGwuqQ0zD0f1vlQai2_Y1N/view?usp=drive_link}{Extension 2}}
\newcommand{\videothree}{\href{https://drive.google.com/file/d/1D78srqTqRDO5W1hP8TYxcIubaeFF5WoJ/view?usp=drive_link}{Extension 3}}
\newcommand{\videofour}{\href{https://drive.google.com/file/d/1WgQE_H0d72eausk3ZEuLQF4hKRCFFfjh/view?usp=drive_link}{Extension 4}}
\newcommand{\videofive}{\href{https://drive.google.com/file/d/1ApppgNE8aeRBZ9ZhFHcNNxaclXsAhbWl/view?usp=drive_link}{Extension 5}}
\newcommand{\videosix}{\href{https://drive.google.com/file/d/1WVwBr_wo7dFbGP7UJ5B-JX7p0KxrRj5j/view?usp=drive_link}{Extension 6}}
\newcommand{\videoExpCoffee}{\href{https://drive.google.com/file/d/1OIT9nUdoZtUJnrOeV3B0x4pbqoI2iZeF/view?usp=drive_link}{Extension 7}}
\newcommand{\videoOnlineAdapt}{\href{https://drive.google.com/file/d/1Jfnf6FUbWZre6D4Mgueze2RsPZK459YP/view?usp=drive_link}{Extension 8}}
\newcommand{\videolatch}{\href{https://drive.google.com/file/d/1gKR2h1dgmFlyd_DP98GA2mn5-gYh04vw/view?usp=drive_link}{Extension 9}}
\newcommand{\videodrawer}{\href{https://drive.google.com/file/d/1o_O5iliA0ChdOgvU8HpHky2CkipUVqCm/view?usp=drive_link}{Extension 10}}
\newcommand{\videoFMBOne}{\href{https://drive.google.com/file/d/1jqqxxp8BudOtbj8cyczovz5aPRclSZE_/view?usp=drive_link}{Extension 11}}
\newcommand{\videoFMBTwo}{\href{https://drive.google.com/file/d/1u4L3yOrLNKvhQgjW1Evy_YJUT-MBUwT1/view?usp=drive_link}{Extension 12}}
\newcommand{\videoFB}{\href{https://drive.google.com/file/d/1zNVonhG-xORD9VWz-XBTVjD0snL9Vxjm/view?usp=drive_link}{Extension 13}}
\newcommand{\videoVSAug}{\href{https://drive.google.com/file/d/15NOw2EB92TA77P1KqCp7i3iFrxv1G2oN/view?usp=drive_link}{Extension 14}}

\title{\LARGE \bf From a Single Demonstration to a General Policy \\ for Contact-Rich Manipulation}

\begin{document}

\author{Xing Li$^{1,2,3}$  \quad\quad\quad  Oliver Brock$^{1,2,3}$
	\thanks{$^{1}$ Robotics and Biology Laboratory, Technische Universität Berlin}
	\thanks{$^{2}$ Science of Intelligence, Research Cluster of Excellence, Berlin}
	\thanks{$^{3}$ Robotics Institute Germany}
	\thanks{We gratefully acknowledge financial support by the Deutsche Forschungsgemeinschaft (DFG, German Research Foundation) under Germany’s Excellence Strategy - EXC 2002/1 ”Science of Intelligence” – project number 390523135. This work has been partially supported by the German Federal Ministry of Research, Technology and Space (BMFTR) under the Robotics Institute Germany (RIG) – grant number 16ME1000.}
}

\makeatletter
\let\@oldmaketitle\@maketitle%
\renewcommand{\@maketitle}{\@oldmaketitle%
	\begin{center}
		\centering
		\vspace{2mm}
		\includegraphics[width=1\linewidth]{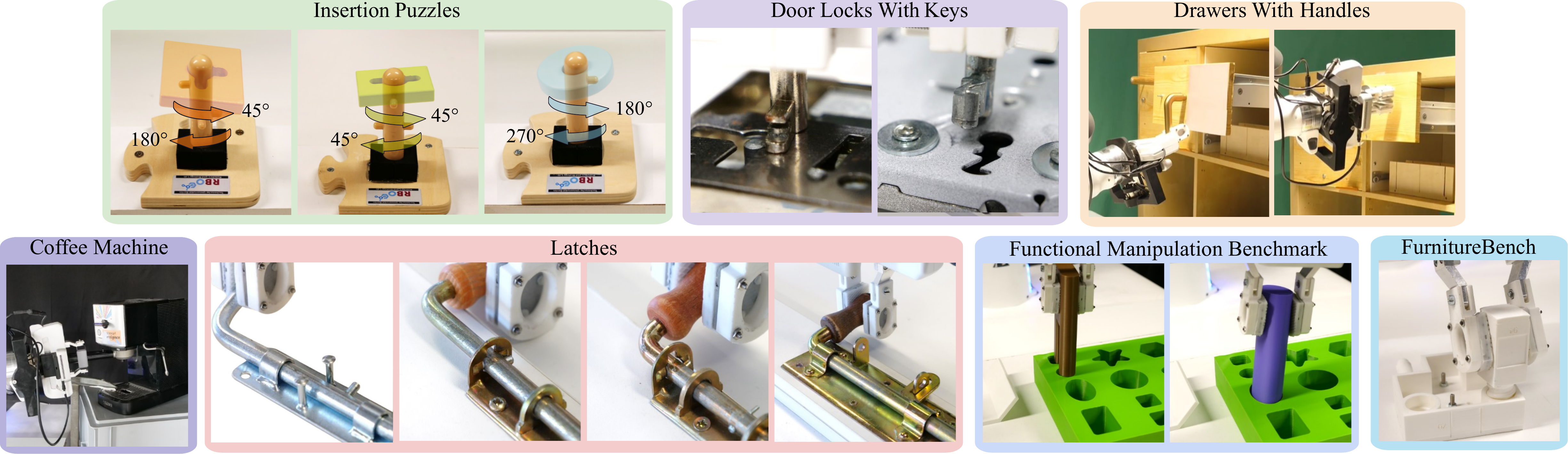}
		\captionof{figure}{Our LfD approach generalizes from a single demonstration and a few corrections more strongly than other approaches in the literature. This generalization manifests itself for a diverse set of multi-stage contact-rich manipulation tasks, including \href{https://drive.google.com/file/d/13yr67N18D8QGwuqQ0zD0f1vlQai2_Y1N/view?usp=sharing}{insertion puzzles},        \href{https://drive.google.com/file/d/1ApppgNE8aeRBZ9ZhFHcNNxaclXsAhbWl/view?usp=drive_link}{door locks with keys},
		\href{https://drive.google.com/file/d/1OIT9nUdoZtUJnrOeV3B0x4pbqoI2iZeF/view?usp=drive_link}{coffee machine},
		\href{https://drive.google.com/file/d/1o_O5iliA0ChdOgvU8HpHky2CkipUVqCm/view?usp=drive_link}{drawers with handles},
		\href{https://drive.google.com/file/d/1gKR2h1dgmFlyd_DP98GA2mn5-gYh04vw/view?usp=drive_link}{latch locks},
		\href{https://drive.google.com/file/d/1jqqxxp8BudOtbj8cyczovz5aPRclSZE_/view?usp=drive_link}{Functional Manipulation Bench}, and
		\href{https://drive.google.com/file/d/1zNVonhG-xORD9VWz-XBTVjD0snL9Vxjm/view?usp=drive_link}{FurnitureBench}. Key to such strong generalization is the use of environmental constraints as inductive biases.}
		\label{fig:fullObjects}
		\vspace{-3mm}
	\end{center}
}
\makeatother

\maketitle
\pagestyle{empty}

\begin{abstract}
	We present a Learning from Demonstration (LfD) framework that achieves one-shot generalization in multi-stage, contact-rich manipulation tasks. Central to our approach is the utilization of environmental constraints as the inductive bias. By representing a demonstration as a sequence of behaviors that exploit environmental constraints, the robot separates task-general structure---the constraint types and their transitions---from instance-specific details such as exact demonstration trajectories, poses, and local geometries. Our four-stage pipeline builds a complete policy on this representation: the robot first abstracts a single demonstration into environmental-constraint primitives, then disambiguates them through self-guided exploration, next assimilates targeted human corrections that handle out-of-distribution variations, and finally recovers the abstracted-away details online through compliant interaction. Because the resulting policy follows constraints rather than mimics trajectories, it generalizes across object poses, local geometries, and unmodeled contact dynamics. We validate our approach on seven real-world multi-stage contact-rich manipulation tasks and achieve over 90\% success. These extensive experimental results establish environmental constraints as fundamental building blocks for efficient generalization in learning from demonstration.
\end{abstract}

\section{Introduction}

Learning from Demonstration (LfD) approaches compute policies from human demonstrations of task executions~\cite{billard2008survey}. The main objective in LfD is for these policies to be as general as possible, i.e., to successfully execute under substantial variations of the task instance. Achieving such generality from as few demonstrations as possible remains a fundamental challenge.

In this paper, we introduce an LfD method that generalizes from one human demonstration to unseen task variations. We test it on seven multi-stage, contact-rich manipulation tasks (see Fig.~\ref{fig:fullObjects}). Compared to prior work, it generalizes better while using only a single demonstration instead of hundreds or thousands.

The proposed approach achieves such strong generalization because it carefully reasons about the structure of the problem, leverages insights about this structure, and uses them as inductive biases for generating and representing the produced policy. As we will see, the most important of these biases are based on environmental constraints (ECs)~\cite{eppner2015exploitation}. The general policy is then determined from a single demonstration based on four conceptual components, each described in more detail in Section~\ref{sec:methodology}.

First, information from the human demonstration is separated into a general part, which is assumed to be shared by all task instantiations, and an instance-specific part. Only the general part is used to produce the general policy. The required separation is achieved based on the inductive bias of ECs. These constraints capture general interaction patterns between the robot and its environment.

Second, the yet incomplete general policy is augmented by the robot via exploring local contact information along the general policy. This complements our understanding of ECs present in the task beyond what could be extracted from the demonstration. The result is an increase in the generality of the policy relative to the first step.

Third, so-called corrective demonstrations further increase the generality of the produced policy. Should the execution of the existing policy fail, more information from the human is required. The human kinesthetically corrects the failure, adding additional hard-to-model details to the policy, further increasing its generality.

Fourth, and finally, the robot fills in the details that vary across object instances using sensor feedback obtained from the interaction with ECs at run time. This step takes the general policy and makes it instance-specific again, leading to the successful generalization of the general policy.

Overall, our method produces general policies by separating out the general information from a demonstration, adding necessary details through augmentation and corrections, and adapting the policy during execution using sensory feedback. Through extensive real-world experiments, we demonstrate that this way of building policies from demonstrations leads to previously unmatched generality.

\section{Related Work}

\subsection{Environmental Constraint Exploitation}
Since our main contribution lies in leveraging ECs to achieve generalization in LfD, we first review how prior works use ECs in robot manipulation. Environmental Constraint Exploitation (ECE) refers to behaviors that exploit useful features of the environment to facilitate manipulation. This concept originates from studies showing that humans deliberately exploit contact to simplify grasping~\cite{kazemi2014human, eppner2015exploitation} and assembly tasks~\cite{klingbeil2017experimental}. While human studies have confirmed the value of ECE, its relevance to robotic manipulation has been recognized for decades. As early as the 1980s, Lozano-Pérez et al. showed that exploiting contact through compliant motions contributes to robustness under uncertainty in assembly tasks~\cite{lozano1984automatic}. Since then, contact exploitation has been applied across many areas, including grasping~\cite{eppner2015planning, bonilla2014grasping, bimbo2019exploiting, pall2021analysis}, planning~\cite{sieverling_interleaving_2017, toussaint2014dual, toussaint2018differentiable, mao2023learning}, robot co-design~\cite{deimel2017automated, vaish2024co}, end-effector design~\cite{catalano2014adaptive, odhner2014compliant, deimel2016novel, puhlmann2022rbo}, extrinsic manipulation~\cite{erdmann2002exploration, hou2020manipulation, ren2024collision}, and in-hand manipulation~\cite{dafle2014extrinsic, bhatt2022surprisingly, patidar2023hand}. We incorporate ECE into our LfD framework by representing demonstrations as sequences of ECE behaviors.

\subsection{One-Shot Imitation Learning}

Recent advances in vision and language foundation models have driven growing interest in end-to-end imitation learning, where observations from demonstrations are mapped directly to actions. Although these methods have been successfully applied to a variety of manipulation tasks, they are typically data-intensive and often require hundreds or thousands of demonstrations per task~\cite{rt1, rt2, o2024open, khazatsky2024droid, bjorck2025gr00t, team2025gemini, lbm2025careful, bu2025agibot}.

An alternative direction is one-shot imitation learning, which aims to learn a task from a single demonstration. Many approaches in this category formulate LfD as a pose estimation problem using object-centric representations such as object keypoints~\cite{xue2022useek, cai2024visual, di2024dinobot, allu2025hrt1, ding2025tref, chen2025backbone, wichitwechkarn2025annotation, keypointLLM2} or poses~\cite{lee2020guided, vitiello2023one, heppert2024ditto, allu2025hrt1}. A demonstrated behavior can then be reproduced by executing the trajectory defined relative to the task object. Since object-centric representations abstract away task-irrelevant details like visual descriptors and object poses, these methods generalize well to such variations. When combined with advanced vision techniques, recent work has shown that this learning paradigm allows the acquisition of thousands of manipulation behaviors from only tens of hours of demonstration data~\cite{kamil2025}.

However, within this object-centric, trajectory-replay formulation, actions are typically represented as explicit Cartesian trajectories relative to the object~\cite{lee2020guided, vitiello2023one, heppert2024ditto}. In the context of contact-rich manipulation, this representation inherently overfits to specific object geometries; a policy learned from a particular latch or door handle rarely generalizes to geometrically distinct instances. Furthermore, tracking Cartesian trajectories is highly susceptible to failure due to unmodeled contact dynamics, such as slippage. To address these limitations, we introduce a novel action representation based on ECs. Building on this representation, our approach achieves superior generalization across varying object geometries and significantly enhances robustness against positional uncertainties.

\subsection{Learning With Manipulation Primitives}

Designing manipulation primitives enables the injection of inductive biases that capture task-relevant structure and thereby improve generalization to task-irrelevant variations~\cite{felip2013manipulation, kroemer2021review}. This idea has been used to learn contact-rich manipulation skills such as object alignment~\cite{skubic2000acquiring}, part assembly~\cite{6DlinearMotions, huLfD6DAssemble}, and long-horizon extrinsic manipulation~\cite{wu2024one}. Our approach follows this direction by representing a demonstration as a sequence of parameterized primitives that exploit ECs.

The most notable distinction between our framework and prior work is that we do not attempt to extract fully specified primitives from demonstration. Rather, we deliberately leave instance-specific parameters under-specified, allowing them to be continuously updated online via sensory feedback. This online adaptation fills in instance-specific details that differ from the demonstration and accounts for uncertainties, leading to strong generalization in real-world manipulation tasks.

\subsection{Learning Geometric Constraints From Demonstrations}

Since our EC primitives are built on geometric constraint models, our work is closely related to research on extracting such constraints from demonstrations~\cite{dang2010robot, niekum2015online, perez2017c, liu2019learning, jain2020learning, coRL2018EC, wake2021learning, liRAL, hegeler2023teaching, gao2022k, huLfD6DAssemble, pfisterer2025helping, mohammadi2024automatic, overbeek2025versatile, overbeek2025identifying}. These approaches typically fit geometric models, e.g., prismatic or revolute joints, to end-effector trajectories. However, this mapping is often ambiguous, as a single trajectory may be consistent with multiple constraint types.

Some prior approaches attempt to address this ambiguity by incorporating force information~\cite{coRL2018EC, overbeek2025versatile}. However, as noted in~\cite{liIROS, suomalainen2022survey}, human demonstrations could lack the necessary contact information, especially as demonstrators get familiar with the task and their motions become more exploitative. Our approach addresses this by augmenting the demonstration with active exploration that generates additional contact information for the disambiguation. While the idea of learning constraints with active exploration has been explored in prior work~\cite{hausman2015active, ortenzi2016kinematics, lin2020contact}, these approaches typically assume that each task is covered by a single constraint. In contrast, we consider multi-stage manipulation tasks involving sequential constraints. We also present a method to compute the most informative exploratory actions to efficiently disambiguate constraint candidates.

\subsection{Interactive Learning From Demonstration}
Finally, we summarize research on interactive learning from demonstration. This line of research has received growing attention as a way to improve policy robustness through human-in-the-loop corrections~\cite{ross2011reduction, coA2}. A critical prerequisite for such correction is failure detection. Existing approaches often rely on out-of-distribution detection~\cite{nemec2020learning, takeuchi2023motion, liu2023modelbased, vanc2025ilesia} or supervised classifiers trained to distinguish between nominal and failure states~\cite{kappler2015data, niekum2015learning}. In contrast, our method detects failures by monitoring contact transitions extracted during a self-guided exploration phase. Deviations from expected transitions signal that a manipulation primitive has terminated incorrectly.

Once a failure is detected, we incorporate human corrections to recover. Different from prior methods that represent corrections as Cartesian trajectories guiding the robot to particular poses~\cite{ross2011reduction, niekum2015learning}, we model corrections as behaviors that bring the robot back to a nominal contact state, from which it can continue to exploit ECs. This idea is conceptually similar to~\cite{xiang2024sc}, where human input guides the robot back to a safe state to avoid premature termination. By expressing corrections at the level of EC-based abstractions rather than low-level actions, we further promote generalization across task variations.

\section{From A Single Demonstration To A General Policy}
\label{sec:methodology}

\begin{figure*}[ht]
	\centering
	\includegraphics[width=.85\linewidth]{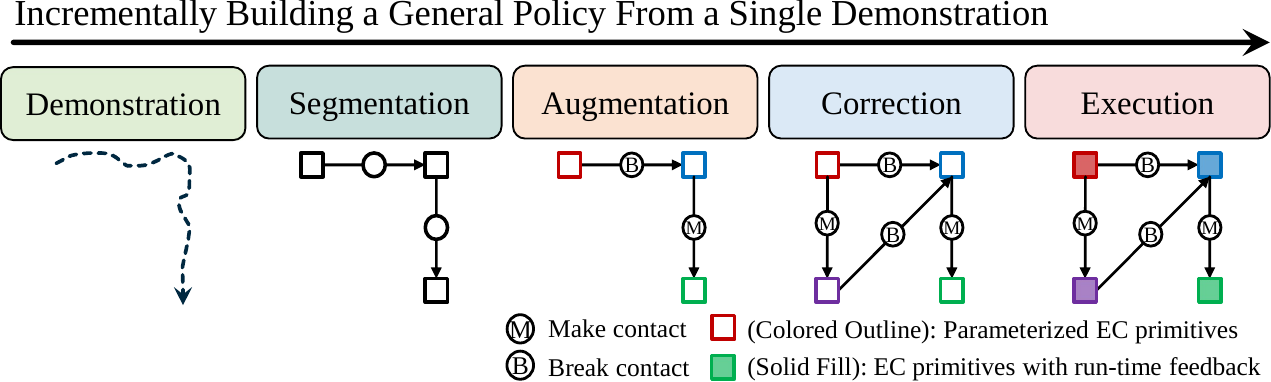}
	\caption{Overview of the four-stage LfD pipeline. A single human demonstration is segmented into a sequence of motion phases, where the exact demonstration actions that are specific to the demonstrated object instance are abstracted using the EC representation. In the next augmentation stage, the robot actively explores the environment around the demonstration to generate additional information of ECs, facilitating the identification of EC primitives (colored outlines) and contact transitions. When the robot encounters unforeseen task variations that are not captured by the EC representation, human corrections are added to expand the policy structure with new EC primitives (purple outline). Finally, during execution, EC primitives (solid fills) leverage sensory feedback to fill in instance-specific details, adapting to the particular instance at run time. Generalization is achieved by separating generalizable information that facilitates learning and leaves the hard-to-model details to be completed via sensory feedback and corrections during execution.}
	\label{fig:lfdPipeline}
\end{figure*}

The core challenge in LfD is generalization. To produce a general policy, we must separate generalizable information of demonstrations from details specific to the demonstration scenario, such as exact object pose and geometry parameters. Without this separation, learning algorithms overfit to individual demonstrations and thus hinder generalization.

Our high-level solution to achieve this separation and generalization is to find the most appropriate inductive bias. In the field of machine learning, such inductive biases are a known precondition for generalization~\cite{mitchell1980need}. We will see in this section that there is a strong inductive bias that naturally enables the derivation of general policy from a single human demonstration, namely, environmental constraints. For clarity, we first present a high-level overview of the methodology that explains why this inductive bias enables generalization and identifies the fundamental components required to derive an EC-based policy. In the next section, we describe how these components are realized and implemented in a concrete LfD approach.

\subsection{Environmental Constraints}

An \textit{environmental constraint} is a feature in the environment that can be used to identify the reference value of a behavior-generating controller~\cite{eppner2015exploitation}. It is important to note that the specific reference value is not specified a priori, but it is extracted from interactions with the environment during execution. Common examples include geometric features in the environment---surfaces, edges, kinematic joints, etc.---which can be used to simplify perception and control by exploiting information obtained through contact. Another set of environmental constraints provides visual rather than haptic references for behavior generation. Such constraints are leveraged in visual servoing. In both cases, the reference value is extracted from sensor data during execution.

This separation supports generalization because ECs arise from an object’s articulation, which encodes its intended functionality. Objects designed for the same function often share the same EC structure; for example, doors, windows, scissors, and drawers exhibit recurring revolute or prismatic constraints. This invariance could be a consequence of human-centered design conventions that standardize how such objects can be used~\cite{norman2013design}. Leveraging this invariance, our approach represents a demonstration as a sequence of EC exploitations.

Consider opening a drawer as an example. First, the robot visually servos toward the handle, whose visual features act as an EC that constrains the end-effector to a grasp pose. Second, after grasping, the robot simply pulls the drawer and lets the drawer’s prismatic joint guide its motion. This motion stops when the drawer is fully opened, signaled by yet another EC, the stop of the drawer.

Therefore, ECs are the basic ``currency'' of a robot's actions in the real world. We will see in what follows that ECs represent the correct inductive bias for LfD. This inductive bias provides a suitable abstraction that avoids overfitting and enables systematic generalization.

\subsection{Separating General From Specific}

Building on ECs, we represent a policy as a hybrid automaton~\cite{egerstedt2000hybridautomaton}, consisting of ECEs (nodes) and transitions (edges). A node represents an ECE controller, implemented as a parameterized manipulation primitive. Each primitive characterizes a specific type of EC through admissible and constrained motion directions and generates motions to follow it accordingly. Nodes are connected via directed edges representing transition conditions. Transitions are triggered by the establishment of a new EC or the removal of a currently exploited one.

This policy structure cleanly separates \emph{general} task information from \emph{instance-specific} details. The \emph{general} part consists of the EC types and their transitions, while the \emph{instance-specific} information, such as the exact Cartesian trajectories or geometric parameters observed in a particular demonstration, is excluded from the policy representation. Instead, these details are recovered during execution, as described later.

Generalization emerges from this separation. For example, different door handles require different rotation angles before opening the door becomes possible, yet the underlying constraint-based behavior in the contact space---rotating until the door becomes pullable---remains the same. By leveraging this insight, our approach generates motion by following ECs rather than mimicking demonstrated trajectories for better generalization.

\subsection{Completing the Environmental Constraint Using Augmentation}

Since our policy is defined in contact space to enable generalization, deriving a good policy depends on correctly identifying ECs and their transitions from demonstrated trajectories. However, a single demonstration often lacks the required information to derive such a policy completely. In particular, when humans are familiar with a task, they tend to act exploitatively, accomplishing the task with minimal interaction with the environment~\cite{liIROS}. As a result, the recorded demonstration contains insufficient interaction (forces and moments) data for the robot to infer the underlying ECs. Moreover, distinct ECs can lead to similar motions~\cite{coRL2018EC}. Consequently, EC identification from demonstrations is often ambiguous.

To resolve this ambiguity, we extend LfD with self-guided exploratory actions. Rather than treating demonstrations as data to be directly imitated, we use them as guidance for autonomous, active interaction with the environment. Specifically, the robot replays a demonstrated trajectory while performing exploratory motions, generating additional constraint information and thereby completing the data required to derive a fully specified EC-based policy. We refer to this process as \emph{augmentation}.

Using augmentation, the robot autonomously completes the partial EC description obtained in previous steps, enabling even broader generalization, even for contact states that were not included in the demonstration data.

\subsection{Corrective Demonstrations}

While augmentation completes the demonstration data in terms of ECs, it does not change the overall structure of the policy. Naturally, a policy based on just one demonstration might fail when it meets unseen changes in task settings, e.g., appearances and local geometric differences, which are not captured by the EC representation. In such cases, corrective demonstrations are required.

Because EC-based behaviors are interpretable, it becomes easy for the policy to identify failures during execution and ask for corrective demonstrations. Moreover, because the policy is compositional, each correction behavior can be incorporated as an additional EC primitive into existing policies, requiring targeted updates rather than extensive retraining. In this way, we incrementally extend policy generality to handle out-of-distribution task variations that are difficult to learn from limited data or to model explicitly.

\subsection{Filling in Instance-Specific Details to Complete the Policy}

After constructing a generalized EC-based policy through augmentation and corrective demonstrations, the instance-specific details previously abstracted away to facilitate learning---such as exact object poses, contact locations, and local geometries---must be recovered to fully parameterize the policy. This happens during execution of the policy based on sensory feedback obtained from interaction with the current task instance.

To this end, we leverage compliant control, which allows the robot to automatically adapt its motion to the encountered ECs. The resulting force and motion feedback is used to update the internal EC model online. This online adaptation shifts the burden of handling instance-specific variations from learning to compliant control, enabling in-context adaptation that supports generalization.

Taken together, the separation of general and instance-specific information, combined with augmentation, corrective demonstrations, and compliant adaptation, forms a complete pipeline for LfD (see Fig.~\ref{fig:lfdPipeline}). As we will demonstrate in the following sections, an LfD approach built following this pipeline achieves substantial generalization in complex, multi-stage contact-rich manipulation tasks.

\section{Implementing LfD for Multi-Stage Contact-Rich Manipulation}

We now describe how the proposed LfD pipeline is implemented to teach a robot manipulation tasks that combine in-contact motions (e.g., insertion and threading) and free-space motions (e.g., grasping and placing).

A key observation is that the interaction patterns within these tasks can be captured by four types of ECs: \emph{translation}, \emph{rotation}, \emph{plane}, and \emph{free-space}. Translation and rotation reflect prismatic and revolute joints common in articulated objects, while plane constraints frequently arise in assembly tasks where objects must slide along surfaces to achieve alignment or insertion. The free-space primitive uses visual feedback to reproduce contact-free motions, such as grasping or placing objects. We model these transitions between EC primitives as discrete events, including \emph{contact-making}, \emph{contact-breaking}, and \emph{gripper} events (i.e., grasp and release).

In the remainder of this section, we first describe demonstration recording and segmentation, then introduce the EC primitives that form the policy. We next present augmentation strategies that support policy construction and corrective demonstrations for policy refinement. Finally, we describe how sensing and compliant control are used during execution to recover the instance-specific details.

\subsection{Demonstration and Segmentation}

We collect a kinesthetic demonstration using a Franka Emika Panda robot~\cite{haddadin2022franka} with a parallel gripper, a wrist-mounted force/torque sensor, and an end-effector–mounted RGB-D camera. Each demonstration waypoint includes the end-effector pose $\mathbf{T}_\mathrm{BE}$, where $\mathrm{B}$ denotes the base frame and $\mathrm{E}$ the end-effector frame, the gripper state $g \in (0,1)$, the measured wrench $F \in \mathbb{R}^6$, and the RGB-D image $\mathrm{I}$.

We segment a demonstration based on multi-modal sensory signals. We first use gripper state changes to segment grasp and release actions. Since gripper states alone do not fully capture constraint transitions, we further analyze force measurements. We compute the force norm and apply Total Variation Denoising (TVD)~\cite{tvd} to smooth the signal while preserving edge information. We then identify rising and falling edges using a \SI{10}{\newton} threshold as segments~\cite{tumHapticExplo}. Segments with an average force below \SI{2}{\newton} are labeled as free-space motions, while others are labeled as in-contact motions.

Since individual in-contact segments may encompass multiple ECs, we further refine these segments using Zero-Velocity-Crossing (ZVC)~\cite{ZVCs}. This method identifies discontinuities in end-effector velocity that typically correspond to transitions between ECs. We avoid over-segmentation by removing any segments where the robot moves less than \SI{2}{\centi\meter}.

After segmenting the demonstration, we have a sequence of motion phases. Each phase will be turned into one of our four EC primitives (Fig.~\ref{fig:ecprimitives}). To keep the explanation clear, we will first describe how these EC primitives work and then explain how the system identifies which one to use for each part of the task.

\begin{figure}[htbp]
	\centering
	\includegraphics[width=\linewidth]{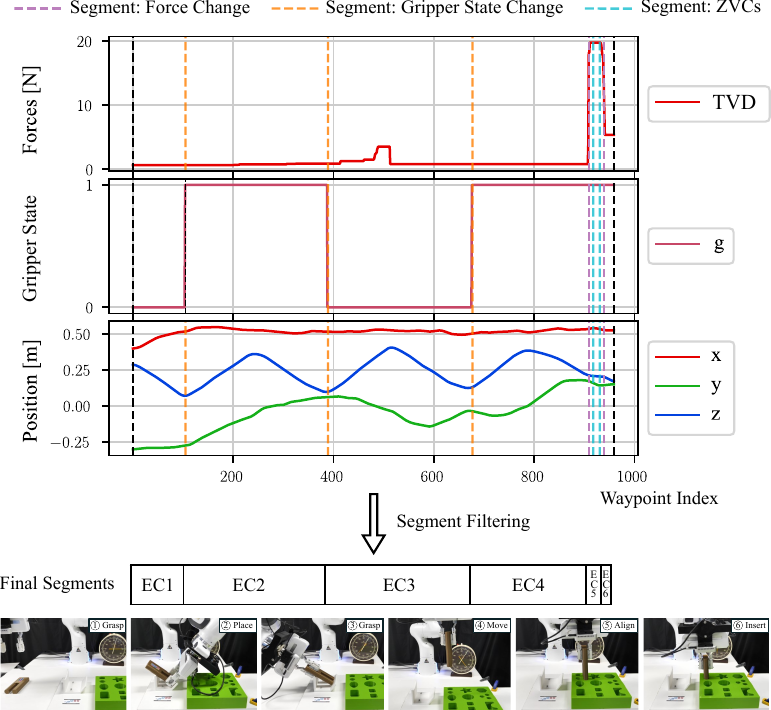}
	\caption{Segmentation result of a demonstration for a single-object insertion task from Functional Manipulation Benchmark~\cite{fmb2023}. The top plots show TVD force signals, gripper states, and end-effector positions, indexed by waypoint of a demonstration. Vertical dashed lines indicate segmentation from force thresholds (purple), gripper state changes (orange), and ZVCs (cyan). We first merge these segments and then remove those whose total displacement is less than \SI{2}{\centi\meter}. The bottom row shows six final segments and snapshots of the robot behavior during each motion phase.}
	\label{fig:fmbsegmentation}
\end{figure}

\subsection{EC Primitives}

\begin{figure*}[ht]
	\centering
	\includegraphics[width=1\linewidth]{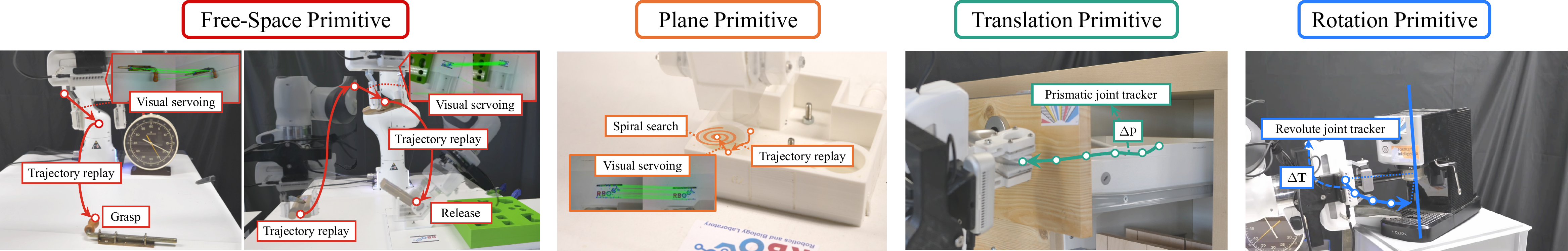}
	\caption{Overview of the four EC primitives. The \textbf{free-space primitive} aligns the end-effector to the target object through visual servoing using a wrist-mounted camera, enabling adaptation to changes in object pose. It then replays the transferred demonstrated trajectory to reproduce contact-free motions. The \textbf{plane primitive} first replays the transferred demonstration trajectory along a surface and then refines its position through visual servoing and spiral search, enabling generalization to uncertain or shifted hole locations. The \textbf{translation and rotation primitives} generate compliant motions that follow prismatic and revolute constraints, respectively, and use observed end-effector movements to update their internal models, allowing generalization to different object geometries. Together, these four EC primitives generate motions that generalize across unseen \emph{object poses} and \emph{geometries}.}
	\label{fig:ecprimitives}
\end{figure*}

\textbf{Free-Space Primitive.}
This primitive is designed to replicate contact-free motions like grasping an object. We implement this primitive based on Trajectory Transfer~\cite{vitiello2023one}, where the robot aligns its current visual input with a reference image from the demonstration via SIFT-based visual servoing~\cite{servo1}. The visual alignment places the end-effector in the same pose relative to the object as in the demonstration. This allows the subsequent trajectory and EC primitives to execute in an object-centric frame, thereby enabling generalization across different object poses.

Methods based on Trajectory Transfer typically utilize the first frame of the demonstration as the visual reference. However, in multi-stage tasks, these methods might fail after a grasp because the wrist-mounted camera is too close to the object to extract enough features for subsequent visual alignment. To mitigate this, we propose a selection strategy that identifies the reference image at the waypoint of maximum elevation in the task space (i.e., the peak $z$-coordinate relative to the initial pose). Our revised procedure involves: (1) replaying the trajectory up to that waypoint, (2) aligning the current view to the reference image, (3) transforming and executing the rest of the trajectory in the aligned end-effector pose, and (4) executing the gripper actions according to the demonstrated gripper states. This design allows the free-space primitive to handle repeated grasping and repositioning (see Fig.~\ref{fig:ecprimitives}).

Taken together, the free-space primitive is parameterized by a reference image, an end-effector trajectory, and gripper states. A SIFT-based visual servoing controller is used to minimize feature errors between current and reference views~\cite{VS3D_2}.

\textbf{Plane Primitive.}
The plane primitive reproduces motions along flat surfaces, such as sliding or guiding objects into slots. To ensure the policy remains independent of absolute Cartesian coordinates, we fit a planar model to the segmented demonstration trajectory to extract the surface normal vector $\hat{\mathrm{n}} \in \mathbb{R}^3$. During execution, the specific plane location is instantiated by the end-effector pose at the moment of making contact with the surface. The primitive then transfers the demonstrated trajectory to the local end-effector frame and replays it while maintaining contact through the Cartesian compliant controller (Sec.~\ref{sec:control}).

After replaying the trajectory, the robot performs SIFT-based visual servoing toward the reference image from the last waypoint of the motion phase and stops when a displacement threshold is reached. It then executes a spiral search on the surface~\cite{spiralsearch}. These two steps help generalize to uncertain or shifted hole locations.

A plane primitive is parameterized by an end-effector trajectory and a reference image from the segmented motion phase. It terminates upon detecting: a tangential force exceeding a threshold (making-contact event), a displacement along the surface normal direction exceeding a threshold (breaking-contact event), or no contact is found after finishing the spiral search (non-contact event).

\textbf{Translation Primitive.} The translation primitive models 1-DOF linear motions, such as pulling a drawer or inserting an object into a hole. Like the plane primitive, the translation primitive avoids using a fixed prismatic model from the demonstration, but maintains a prismatic joint tracker that online estimates the admissible motion direction based on observed end-effector displacements. As evidenced in our real-world experiments, this online adaptation allows the robot to generalize to unseen object geometries or inaccurate models.

The prismatic joint tracker is implemented as a particle filter~\cite{liRAL}, where each particle represents a motion direction $\hat{\mathrm{u}}^\mathrm{t} \in \mathbb{R}^3$ parameterized in spherical coordinates $(\theta^{\mathrm{t}}, \phi^{\mathrm{t}})$ (the superscript $\mathrm{t}$ denoting the translational motion type). During the prediction step, particles are perturbed with zero-mean Gaussian noise. We then sample a candidate direction from the particle distribution and execute a compliant motion along it. If the resulting linear motion exceeds \SI{3}{\milli\meter}, the filter increases the weights of particles that are close to the observed motion direction; otherwise, it skips the update, which increases the distribution's variance and thus encourages more exploration in subsequent steps.

The translation primitive is initialized by fitting a line to the segmented position trajectory to obtain $\hat{\mathrm{u}}^\mathrm{t}$. Execution terminates upon detecting a contact change, either when no update step is performed after a maximum number of sampling attempts (making-contact event), or when the robot detects significant displacement in a previously constrained direction (breaking-contact event).

\textbf{Rotation Primitive.}
The rotation primitive reproduces motions around revolute constraints, such as rotating a handle or screwing a cap. The rotation primitive follows the same sample-execute-update process as the translation primitive: it maintains a revolute joint tracker, samples a desired end-effector pose from the current estimate of the revolute joint parameters, executes it with compliant control, and updates the estimate based on the observed motion. This process allows the robot to adapt to local variations in the revolute constraint parameters and thus generalize across unseen scenarios.

The revolute joint tracker is implemented as the Extended Kalman Filter (EKF) presented in~\cite{martin2022coupled}, which maintains a belief over the screw parameters $(\hat{\mathrm{u}}^{\mathrm{r}},\: \mathrm{p}^{\mathrm{r}},\: q^{\mathrm{r}},\: \dot{q}^{\mathrm{r}})$, where $\hat{\mathrm{u}}^{\mathrm{r}} \in \mathbb{R}^3$ is the rotation axis (with the superscript $\mathrm{r}$ denoting the rotational motion type), $\mathrm{p}^{\mathrm{r}} \in \mathbb{R}^3$ is a point on the axis, $q^{\mathrm{r}} \in \mathbb{R}$ is the rotation amount, and $\dot{q}^{\mathrm{r}} \in \mathbb{R}$ is its angular velocity.

At each step, the EKF propagates $q$ using $\dot{q}^{\mathrm{r}}$ and injects process noise into the state. A rotation axis is then sampled from the EKF belief to compute the next motion command $z \in se(3)$ using the measurement model:
\[
z =
\begin{pmatrix}
	q^{\mathrm{r}} \; \hat{\mathrm{u}}^{\mathrm{r}} \\
	- q^{\mathrm{r}} \; \hat{\mathrm{u}}^{\mathrm{r}} \times \mathrm{p}^{\mathrm{r}}
\end{pmatrix},
\]
where the first term corresponds to the angular motion and the second term to the linear motion. The desired end-effector pose is then computed as $\mathbf{T}_{\mathrm{BE}} = \exp(z^\wedge) \; \mathbf{T}_{\mathrm{BE}_{\mathrm{init}}}$, where $\exp(\cdot^\wedge)$ denotes the matrix exponential map and $\mathbf{T}_{\mathrm{BE}_{\mathrm{init}}}$ is the initial end-effector pose at the start of the primitive. Similar to the prismatic joint tracker, the EKF only performs an update when the observed linear motion exceeds \SI{1}{\centi\meter} or the angular motion exceeds \SI{2}{\degree}; otherwise, the update is skipped. Skipping the update increases the uncertainty of the states and encourages the robot to explore different rotation axes in the subsequent sampling steps. This prevents the robot from getting stuck due to inaccurate model parameters.

The rotation primitive is initialized using the screw parameters $\hat{\mathrm{u}}^{\mathrm{r}}$ and $\mathrm{p}^{\mathrm{r}}$ obtained by fitting a revolute constraint model to the segmented demonstration trajectory. The termination conditions are the same as for the translation primitive.

\subsection{Augmentation}

After segmentation, we must instantiate an EC primitive for each motion phase by assigning it to the correct EC type. This assignment is non-trivial for in-contact motion phases due to inherent kinematic ambiguities. For example, an end-effector trajectory following a prismatic or revolute joint may also satisfy a plane constraint that happens to be a line or circle. It is also difficult to determine whether a primitive terminates and transitions due to making a new contact or breaking an existing one, since noisy force measurements can obscure contact-changing events.

To address these challenges, we introduce a \emph{contact-based augmentation} strategy in which the robot replays the demonstration trajectory while performing exploratory probing actions to actively verify the candidate constraint directions and contact transitions. Additionally, we propose a \emph{vision-based augmentation} strategy to improve visual alignment in free-space primitives by enhancing image features.

\subsubsection{Contact-Based Augmentation}

As described earlier, we need to assign an EC type to a segmented motion phase. To do so, we first classify motion phases as contact-free or in-contact using force thresholds. A free-space primitive is directly assigned to contact-free motions. For in-contact motions, we initially distinguish translation and rotation ECs using an angular displacement threshold of \SI{30}{\degree}. To resolve remaining geometric ambiguities, we replay the demonstration and control the robot to probe a direction $\hat{\mathrm{d}} \in \mathbb{R}^3$ at the midpoint of each in-contact motion:

\[
\mathrm{\hat{d}}=
\begin{dcases}
	\mathrm{\hat{u}}^\mathrm{t} \times \mathrm{\hat{n}}, \; \text{translation}
	\\
	(\mathbf{I} - \mathrm{\hat{u}^\mathrm{r}} \; \mathrm{\hat{u}^{\mathrm{r}, T}}) \; (\mathrm{p} - \mathrm{p}^\mathrm{r}), \; \text{rotation}
	\\
	\mathrm{\hat{n}}, \; \text{plane}
\end{dcases}
\]

where $\mathrm{p}\in \mathbb{R}^3$ is the end-effector position, $\mathrm{\hat{n}}$ is the surface normal direction, $\mathrm{\hat{u}}^\mathrm{t}$ and $\mathrm{\hat{u}}^\mathrm{r}$ are the translation and rotation axes, and $\mathrm{p}^\mathrm{r} \in \mathbb{R}^3$ is a point of the rotation axis. These parameters are computed for each motion phase by fitting geometric constraint models to the segmented trajectory using the method of~\cite{coRL2018EC}.

If the robot observes a constraint along the direction $(\mathbf{I} - \hat{\mathrm{u}}^\mathrm{r} \; \mathrm{\hat{u}^{\mathrm{r}, T}}) \; (\mathrm{p} - \mathrm{p}^\mathrm{r})$, the motion phase is classified as a rotation EC, since this direction would remain unconstrained for a plane EC. Similarly, if no constraint is detected along $\hat{\mathrm{u}}^{\mathrm{t}} \times \hat{\mathrm{n}}$, the motion phase corresponds to a plane EC, as this direction is constrained by a translation EC.

Following our prior work~\cite{liIROS}, we further verify contact-changing events between consecutive motion phases. After disambiguating the EC type of a motion phase, the robot compliantly executes the trajectory of the next motion phase. If the motion is blocked by a constraint, we classify the transition as a breaking-contact event; otherwise, we label it as a making-contact event. Examples of this contact-based augmentation strategy are shown in \videoone~and \videofour.

Overall, the contact-based augmentation strategy actively verifies constraint directions around the demonstration trajectory, generating additional contact information that supports the recognition of EC types and contact transitions.

\subsubsection{Vision-Based Augmentation}

The performance of visual alignment within the free-space primitive depends heavily on the viewpoint invariance of SIFT features. To improve alignment performance, we introduce a vision-based augmentation that prunes SIFT features that are prone to causing mismatches under changing views.

Given a free-space primitive, we first reset the object to its initial demonstration pose. The robot then selects a reference image, extracts SIFT features, and projects them into 3D using the depth data. Next, it executes a scripted motion that actively varies the camera viewpoint. For each newly captured image, we match the observed SIFT features to the reference features using FLANN-based matching~\cite{opencv_library} and project the matched keypoints into 3D.

Since the target object remains static, the quality of each feature match can be evaluated by measuring the 3D Euclidean distance between the matched feature and its original position in the reference image. Matches with a displacement smaller than \SI{2}{\centi\meter} are considered valid. For each reference SIFT feature, a matching score is computed as the normalized count of valid matches observed across all viewpoints. Features with scores below 0.2 are labeled as viewpoint-inconsistent and removed.

\begin{figure}[t]
	\centering
	\includegraphics[width=1\linewidth]{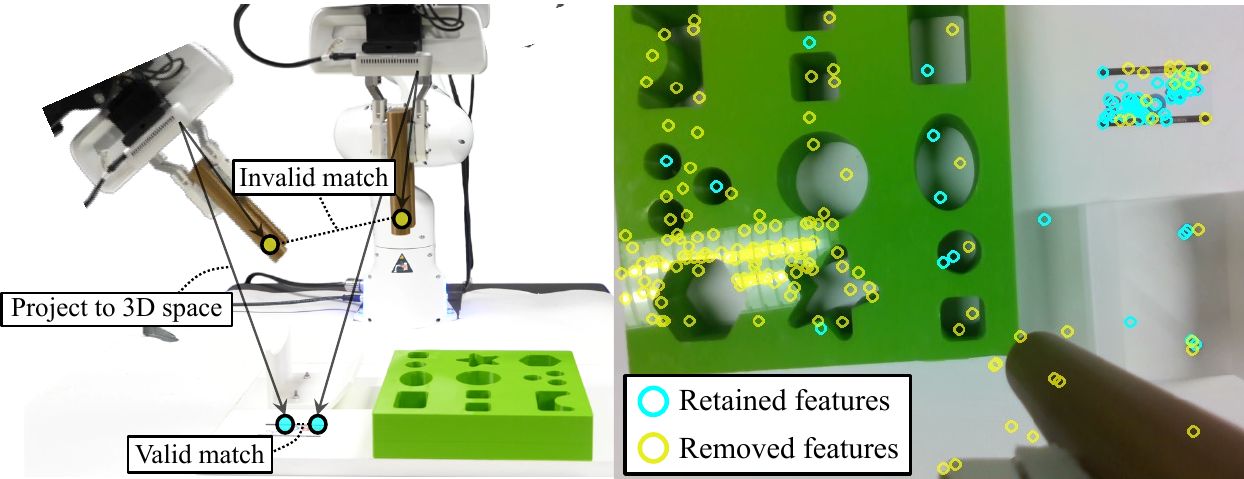}
	\caption{Vision-based augmentation to filter unstable features. By actively changing the robot’s viewpoint and evaluating feature matching performance (left), the vision-based augmentation strategy filters out features that are not consistent under viewpoint changes (right). This includes features caused by light reflections on the insertion board surface, which are prone to mismatching and sensitive to lighting variation. It also filters out features on the inserted object. Because these features move with the robot, they are inconsistent with changing viewpoints and would introduce errors to the visual alignment process.}
	\label{fig:visionAugmentation}
\end{figure}

As exemplified in Fig.~\ref{fig:visionAugmentation} and \videoVSAug, the proposed vision-based augmentation strategy effectively filters out features prone to mismatching. The results presented in Appendix~\ref{appendix:visualAugmentation} further demonstrate that the enhanced visual alignment improves the generalization to lighting changes.

\subsection{Policy Refinement With Human Corrections}

Following augmentation, the resulting policy comprises a sequence of EC primitives linked by discrete contact-state transitions and gripper events. Although this policy captures the general task structure, executions may still fail when the policy encounters out-of-distribution variations that are not modeled by the EC representation. However, the discrete nature of the policy inherently facilitates failure detection. By continuously monitoring transition events, the policy can evaluate whether an EC primitive terminates nominally. For instance, a missed insertion manifests as the absence of an expected breaking-contact event, whereas a lost grasp yields an unexpected gripper state.

Once an execution failure is detected, the user can provide a corrective demonstration. Specifically, we implement two ways to incorporate human corrections. If a free-space primitive ends incorrectly (e.g., a lost grasp), the user provides a new demonstration that replaces the reference image and associated trajectory, allowing the primitive to adapt to the new scenario, such as changes in object appearance. For a contact-based primitive, the user demonstrates a recovery trajectory that brings the robot to a state where the original failed EC can resume execution. A new primitive of the same EC type is instantiated and linked to the failed EC through the newly observed contact transition. When this corrective primitive is triggered, the robot first executes the recovery trajectory and then continues with the associated EC-based motions. An example of the correction process is shown in~\videosix$\,$.

\subsection{Sequencing EC Primitives as Funnels}
\label{sec:sequencing}

During the execution of an EC-based policy, we enhance robust transitions between EC primitives by intentionally establishing or maintaining contact with the environment. This is grounded in the sequential composition of funnels~\cite{burridge1999sequential}, where each EC primitive acts as a funnel that guides the system into the specific contact state required by the subsequent primitive.

\begin{figure}[t]
	\centering
	\includegraphics[width=1\linewidth]{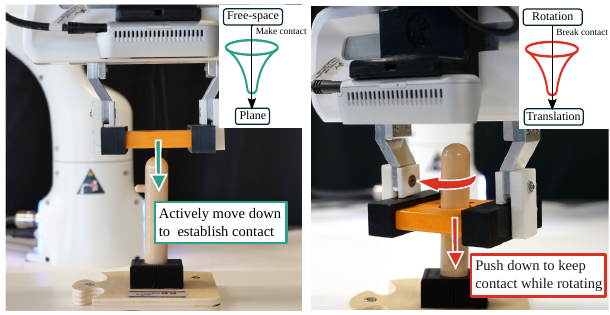}
	\caption{Illustration of our funneling strategies in the puzzle insertion task. \textbf{Left}: After a free-space primitive, the robot stops just short of contact. Transitioning to the plane primitive from this state would erroneously trigger a breaking-contact event, leading to premature state transition and subsequent manipulation failure. Our funneling strategy addresses this by moving along the plane’s constraint direction to establish contact, compensating for positional uncertainties. \textbf{Right}:
	During a rotation primitive, the robot applies a force along the constrained axis (the admissible direction of the subsequent translation primitive) to maintain contact. This simplifies the detection of the desired contact-breaking event.}
	\label{fig:sequenceECEs}
\end{figure}

Specifically, we implement two funneling strategies (see Fig.~\ref{fig:sequenceECEs}). First, when transitioning from a free-space motion to an in-contact primitive, the policy proactively commands a motion along the anticipated constraint direction of the upcoming EC to establish contact. Second, when a breaking-contact transition is expected, the robot deliberately maintains contact by applying force along the admissible direction of the next EC. These two strategies deliberately establish or maintain contact to expose the desired contact feedback during execution, facilitating transitions between EC primitives and improving robustness to uncertainties in pose or geometry.

\subsection{Cartesian Compliant Controller}
\label{sec:control}

An EC-based policy generates behaviors that alternate between free-space and in-contact motions. Free-space motions benefit from high-stiffness control to ensure precise execution, whereas in-contact motions require more compliance to adapt to local geometry. Also, the robot must handle frequent and unpredictable contact transitions. Thus, we need a compliant controller that is adaptive and contact-aware.

We contribute a Cartesian compliant controller which meets these requirements. The key idea is to integrate wrench feedback from a wrist-mounted force/torque sensor into the trajectory generator that modulates the desired equilibrium pose~\cite{jainEPC}.

Given a target pose G in the base frame B, namely $\mathbf{T}_\mathrm{BG}$, we first interpolate the goal in exponential coordinates $\Delta\psi_\mathrm{t} \in se(3)$ subject to an element-wise twist limit $V_\mathrm{m} \in \mathbb{R}^6$ by

\begin{equation}
	\Delta\psi_{t} = \max\left( \min\left( \mathrm{log}(\mathbf{T}^*_\mathrm{BE_{t-1}} \: \mathbf{T}_\mathrm{BG})^{\vee}, V_\mathrm{m} \: \Delta t\right), \; -V_\mathrm{m} \: \Delta t \right)
\end{equation}

where $\Delta t=0.001$ corresponds to a 1kHz control frequency, $\mathbf{T}^*_\mathrm{BE_{t-1}}$ is the desired pose at the last time step $\mathrm{t-1}$, and $\mathrm{log}(\cdot)^{\vee}$ denotes the logarithmic map. We further constrain $\Delta\psi_{t} \in se(3)$ based on the difference between the element-wise absolute value of the observed wrench $\lvert F_{\mathrm{obs, t}}\rvert \in \mathbb{R}^6$ and the predefined wrench limit $F_\mathrm{m} \in \mathbb{R}^6$:

\begin{gather}
	\Delta F_\mathrm{t} = F_\mathrm{m} - \lvert F_{\mathrm{obs, t}}\rvert
	\\
	\Delta\psi^*_\mathrm{t}=\left\{\begin{array}{lr} \tanh(|\Delta F_\mathrm{t}|) \; \Delta\psi_\mathrm{t},
		& \Delta F_{t} > 0
		\\
		\mathrm{sign}\left(F_{\mathrm{obs, t}}\right) \; \tanh(|\Delta F_\mathrm{t}|) \; V_\mathrm{m} \: \Delta t,
		& \text { otherwise }
	\end{array}\right.
	\label{eq:test}
\end{gather}

where the $\tanh(\cdot)$ function acts as a smooth factor. When $\Delta F_\mathrm{t} > 0$ i.e., exceed force limits, $\mathrm{sign(\cdot)}$ will flip the interpolation direction to reduce the observed wrench $F_{\mathrm{obs, t}}$. After obtaining the wrench-limited relative pose $\Delta\psi^*_\mathrm{t} \in se(3)$, we compute the desired end-effector pose by $\mathbf{T}^*_\mathrm{BE_{t}} = \mathbf{T}^*_\mathrm{BE_{t-1}} \: \exp(\Delta\psi^{* \: \wedge}_\mathrm{t})$. We then compute the desired motor torques $\tau_\mathrm{cmd} \in \mathbb{R}^7$ following the operational space formulation~\cite{khatib1987unified}

\begin{equation}
	\tau_\mathrm{cmd} = J^T \; \left(K_\mathrm{p} \; e_\mathrm{pose} - K_\mathrm{d} \; V_\mathrm{obs}\right) + \tau_\mathrm{dyn},
\end{equation}

where $K_\mathrm{p} \in \mathbb{R}^{6\times 6}$ and $K_\mathrm{d}\in \mathbb{R}^{6\times 6}$ are positive symmetric controller gain matrices for stiffness and damping respectively, $e_\mathrm{pose} \in \mathbb{R}^6$ is the error between the observed and desired end-effector pose, $V_\mathrm{obs} \in \mathbb{R}^6$ is the observed end-effector twist, $J^T \in \mathbb{R}^{7\times 6}$ is the Jacobian transpose, $\tau_\mathrm{dyn} \in \mathbb{R}^7$ represents the torques that need to be compensated such as gravity and Coriolis forces.

\begin{figure}
	\includegraphics[width=\linewidth]{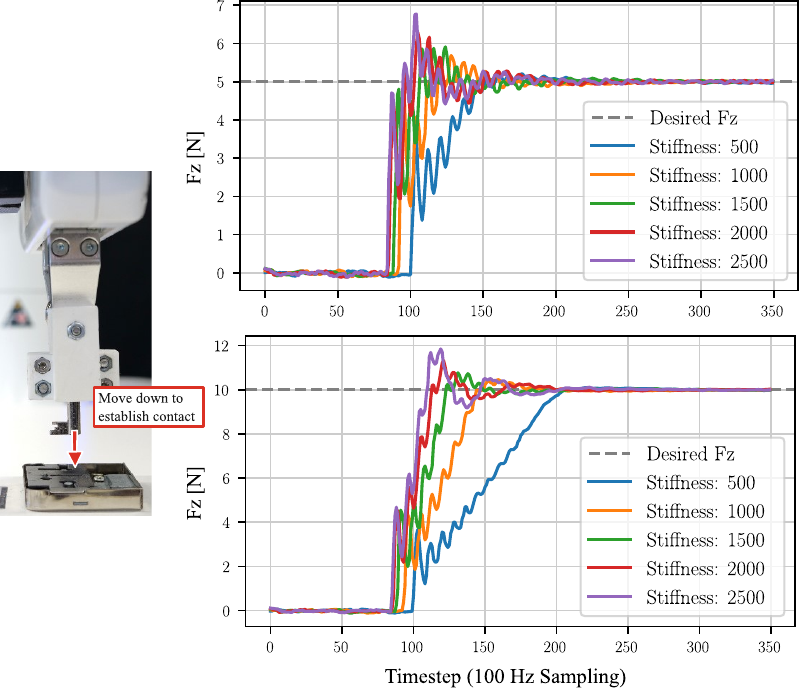}
	\caption{Observed z-axis forces as the robot approaches the lock surface at \SI{2}{\cm/\s}. The controller is evaluated with predefined force limits of \SI{5}{\newton} and \SI{10}{\newton} across stiffness settings from \SI{500}{\newton/\meter} to \SI{2500}{\newton/\meter}. The force profiles show that the controller reliably maintains the target contact force under various stiffness settings, enabling safe interaction and simplifying parameter tuning.}
	\label{fig:RobustToGains}
\end{figure}

We evaluate the controller using a door-lock manipulation task requiring the robot to establish contact with the environment. As illustrated in Fig.~\ref{fig:RobustToGains}, our controller successfully maintains a desired contact force even with high stiffness. Consequently, this controller allows us to directly specify the desired interaction force simply by setting the appropriate wrench limits. This mitigates impact forces during contact transitions, reduces the risk of damaging the manipulated object, and avoids tedious tuning of control gains. These features are essential for implementing EC exploitation behaviors on real robots.

\section{Experiments}
\label{sec:experiments}

We evaluate the generalization capability of our approach on seven real-world contact-rich manipulation tasks spanning insertion, articulated mechanism manipulation, lock operation with a key, and long-horizon assembly (see Fig.~\ref{fig:fullObjects}). These tasks are challenging because they require precise physical interaction and multi-stage sequential execution. Our extensive real-world experiments demonstrate that ECs are the appropriate inductive bias for generalization by (i) abstracting task-general structure based on ECs from instance-specific Cartesian trajectories, (ii) enabling online adaptation to uncertain or imprecise constraint models, and (iii) supporting failure detection and targeted policy refinement through human corrections during execution.

Below, we describe these tasks and their challenges, followed by an outline of the assumptions used in the experiments.

\subsection{Task Overview and Challenges}

\textbf{Insertion Puzzle.} The robot inserts a grasped part into a puzzle board and then performs a sequence of actions so that the part passes through internal pins. While these puzzles share similar visual appearances, each requires a distinct Cartesian trajectory to complete. Another challenge lies in combining precise insertion with large rotations (up to \SI{270}{\degree}).


\textbf{Door Locks with Keys.} The robot inserts a key into a lock and rotates it by \SI{120}{\degree} to change the lock's state. Compared to insertion puzzles, this task is more sensitive to pose uncertainty due to tighter tolerances.


\textbf{Coffee Machine.} The robot inserts a portafilter and rotates it by approximately \SI{45}{\degree} to tighten it. This task requires the robot to operate near its joint limits, which reduces kinematic manipulability and increases execution noise. Additionally, we challenge the system with imprecise EC models.


\textbf{Latches and Drawers.} These tasks require the robot to grasp and operate mechanisms with multiple DOFs. The key challenge is that critical task parameters for deriving Cartesian trajectories, such as the rotation angles of latches and handles, are difficult to infer from visual observations. Even when such trajectories are available, executing them is prone to failure due to unmodeled slippage between the robot’s gripper and the manipulated object during execution.


\textbf{Functional Manipulation Benchmark and FurnitureBench.} These tasks are taken from two recent long-horizon contact-rich manipulation benchmarks and require sequential actions, including grasping, placing, regrasping, inserting, and screwing. Prior work has shown that learning such behaviors end-to-end remains challenging even with large numbers of demonstrations, due to the long-horizon structure and fine-grained contact interactions.

Overall, these tasks involve complex contact dynamics, partial observability of task parameters, variations in object geometry, imprecise geometric parameters inferred from a demonstration, and long-horizon execution under full 6D pose uncertainty. In the following experiments, we demonstrate how using ECs as the inductive bias enables policies to effectively address these challenges.

\subsection{Experimental Assumptions}

To ensure a fair interpretation of the experimental results, we explicitly state the key assumptions and revisit them in Sec.~\ref{sec:limitations}, which also outlines promising directions for relaxing these assumptions.

\begin{itemize}
	\item We assume objects have sufficient and transferable SIFT features for the free-space primitive. We achieve this by using textured image stickers.
	\item In the insertion puzzle, door lock, and coffee machine tasks, we assume the robot begins with a firm grasp of the manipulated object.
	\item We assume that the provided demonstrations are segmented into appropriate motion phases without mis-segmentation.
	\item We assume that objects are placed within the robot’s reachable workspace such that the task can be executed without violating joint limits.
	\item We assume task-relevant fixtures, such as insertion boards and locks, are rigidly mounted on the environment.
	\item We assume task-specific control parameters and contact-transition thresholds can be set a priori (see Table~\ref{tab:parameters}); these constitute minor tuning and do not change the LfD pipeline.
\end{itemize}

\subsection{Insertion Puzzles}
\label{sec:insertionPuzzle}

\begin{figure}
	\centering
	\includegraphics[width=1\linewidth]{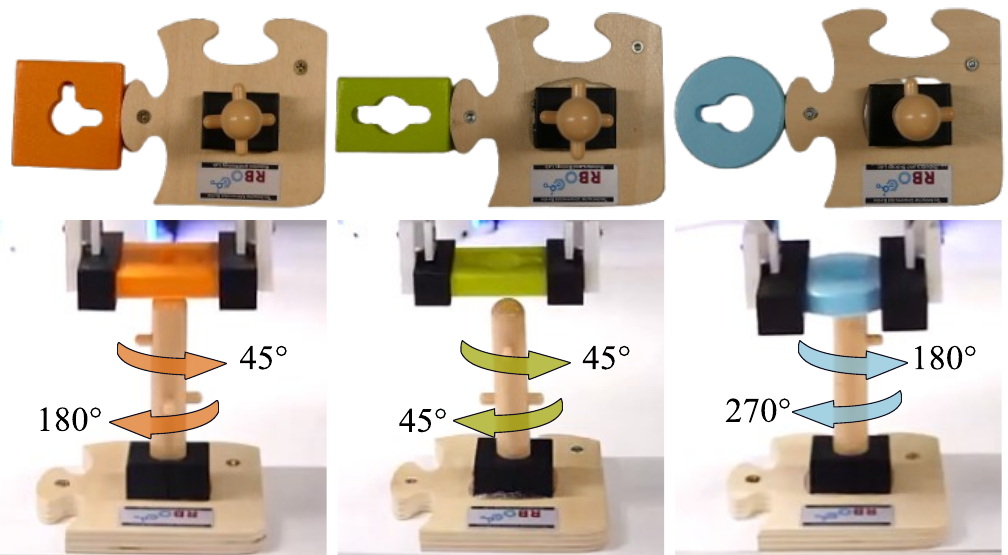}
	\caption{The three insertion puzzles used in our experiments. While each puzzle requires a distinct Cartesian trajectory, their motions in the contact space remain the same.}
	\label{fig:Puzzles}
\end{figure}

We first test our method on insertion puzzle tasks (Fig.~\ref{fig:Puzzles}) to evaluate generalization across object geometries. We record a demonstration using the orange puzzle. Our approach automatically segments this demonstration into seven motion phases and then performs augmentation to disambiguate ECs and contact-changing events. The resulting policy consists of seven EC primitives. The policy construction process is shown in~\videoone. Table~\ref{tab:parameters} lists the control parameters and the thresholds used for detecting contact-changing events.

\begin{table}
	\centering
	\renewcommand{\arraystretch}{1.3} 
	\caption{Control and event thresholds used in the experiments.}
	\label{tab:parameters}
	\resizebox{\linewidth}{!}{%
		\begin{tabular}{lll}
			\hline
			\textbf{Param.} & \textbf{Description} & \textbf{Value} \\
			\hline
			$K_{\mathrm{p, trans}}$ & \makecell[l]{stiffness gains for the translation \\ primitive [\si{\newton/\meter}, \si{\newton\meter/\radian}]}& $\mathrm{diag}[500, 500, 500, 0, 0, 0]$ \\
			\hline
			$K_{\mathrm{p}}$ & \makecell[l]{stiffness gains for other \\ primitives [\si{\newton/\meter}, \si{\newton\meter/\radian}]}& $\mathrm{diag}[1000, 1000, 1000, 200, 200, 200]$ \\
			\hline
			$K_{\mathrm{d}}$ & damping gains [\si{\newton\second/\meter}, \si{\newton\meter\second/\radian}] & $\mathrm{diag}[10, 10, 10, 5, 5, 5]$ \\
			\hline
			$V_\mathrm{m}$ & twist limit [\si{\meter/s}, \si{\radian/s}] & [0.03, 0.03, 0.03, 0.3, 0.3, 0.3] \\
			\hline
			$F_\mathrm{m}$ & wrench limit [\si{\newton}, \si{\newton\meter}] & [5, 5, 5, 2, 2, 2] \\
			\hline
			$r$ & radius of spiral search trajectory & \SI{3}{\cm} \\
			\hline
			$\mathrm{MCE}^\mathrm{p}$ & \makecell[l]{force threshold \\ that triggers making-contact event \\ for a plane primitive} & \SI{5}{\N} \\
			\hline
			\makecell[l]{$\mathrm{MCE}^\mathrm{t}$ and \\ $\mathrm{MCE}^\mathrm{r}$} & \makecell[l]{maximum sampling attempts \\ for a making-contact event \\ in translation and rotation primitives} & 30 \\
			\hline
			$\mathrm{BCE}$ & \makecell[l]{displacement threshold for \\ a breaking-contact event} & \SI{0.8}{\cm} \\
			\hline
		\end{tabular}
	}
\end{table}

We compare our method to five baselines with different policy representations. BC~\cite{BC2} and VINN~\cite{VINN} take wrist camera images as input and output delta end-effector poses. ACT~\cite{ACT} and DP~\cite{DP} use images from a wrist camera and a static camera along with proprioception to predict joint position trajectories. Finally, Trajectory Transfer~\cite{vitiello2023one} performs visual alignment using the initial image from a demonstration and then replays an adapted Cartesian trajectory. We record 30 demonstrations for the end-to-end learning methods (BC, VINN, ACT, and DP), and 1 demonstration for Trajectory Transfer.

\begin{figure}
	\centering
	\includegraphics[width=1\linewidth]{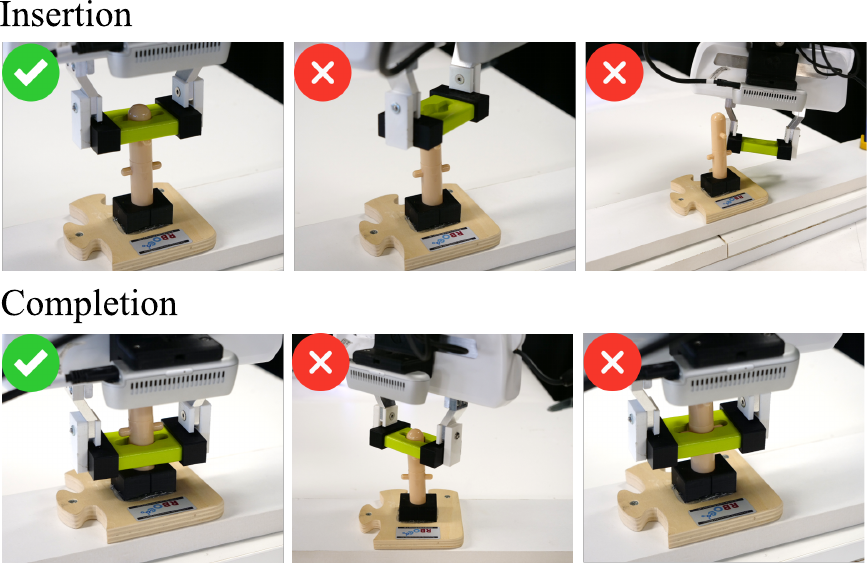}
	\caption{Illustration of success and failure modes in the puzzle insertion task, divided into two stages: \emph{Insertion} (top) and \emph{Completion} (bottom). A trial is considered a successful
	\emph{insertion} if the part is correctly placed into the puzzle board. Failure modes in this stage include misalignment or prematurely lowering the part. In the \emph{completion} stage, success requires guiding the part through the internal pins via two sequential and precise rotations. Failures at this stage include incorrect rotation due to perceptual aliasing and incomplete insertion resulting from perceptual and control inaccuracies.}
	\label{fig:SusccessFailureModes}
\end{figure}

We run 10 trials per method and report success rates. The puzzle board is rigidly fixed on the table. Each trial is initialized from a nominal end-effector pose, subject to uniformly sampled rotational perturbations of $U(-15^\circ, 15^\circ)$ across all three axes. This requires the robot to generate actions in full 6D space. A trial counts as a successful \emph{insertion} if the part is inserted on the top of the board, and a successful \emph{completion} if it passes through the lowest pins (see Fig.~\ref{fig:SusccessFailureModes}).

\begin{figure*}[ht]
	\centering
	\includegraphics[width=1\linewidth]{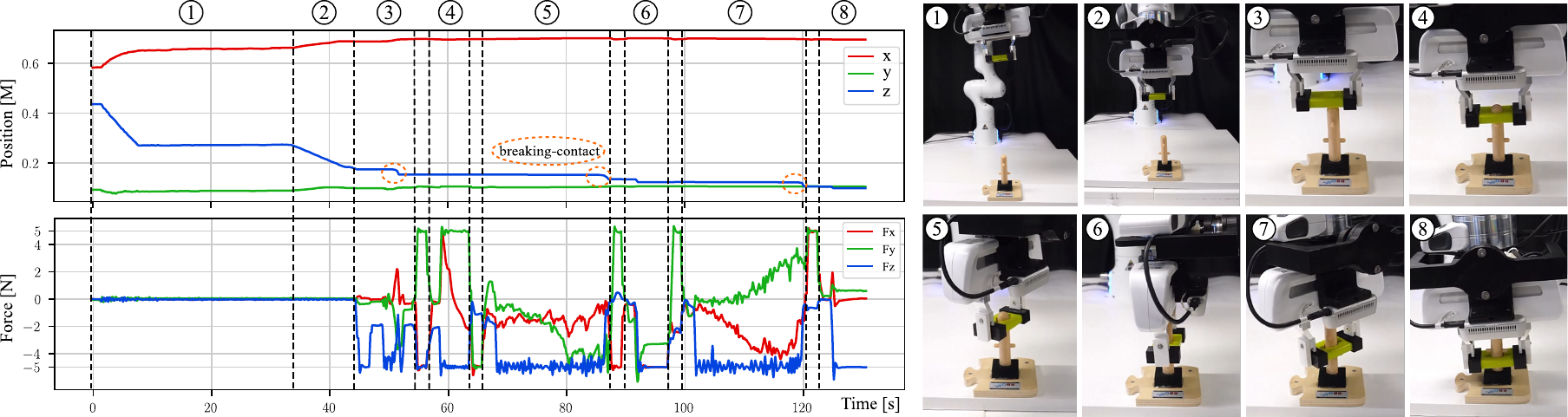}
	\caption{Solving a green puzzle by following a sequence of ECs. \circlednum{1} The robot first visually aligns the end-effector to the puzzle and transfers the demonstration trajectory to the new pose. \circlednum{2} It replays the trajectory and purposefully pushes down until making contact with the puzzle. \circlednum{3} It then executes the plane primitive that replays the transferred trajectory on the plane, followed by a visual servoing and a spiral search action to reduce positional uncertainties, terminating upon successful insertion (a breaking-contact event). \circlednum{4} The robot then pushes the part down by following the prismatic constraint until it makes contact with the pins of the insertion board. \circlednum{5} It rotates the part while pushing, breaking contact again to terminate the rotation primitive. \circlednum{6}-\circlednum{7}-\circlednum{8} The robot repeats the translation and rotation motion until the grasped part is completely fitted into the puzzle. The force plot shows that the robot generates sequential compliant motions sequenced by making or breaking contact events, rather than following absolute positional trajectories, which leads to generalization. It also shows that our compliant controller is able to effectively regulate the force within the predefined threshold (\SI{5}{\N} see Table~\ref{tab:parameters}) during the manipulation, ensuring safe forceful interaction.}
	\label{fig:PuzzleOpeningProcess}
\end{figure*}

Table~\ref{table:Puzzle0Degree} summarizes the results. BC often rotates the end-effector before contacting the board, leading to out-of-distribution visual observations and causing compounding errors. VINN reduces compounding errors by retrieving actions from demonstrations, but it fails due to poor action transitions---it consistently executes the push-down action too early during the approach, leading to failure.

ACT and DP perform better than BC and VINN, achieving insertion-stage success rates of 50\% and 60\%, respectively. This performance gain is attributed to their utilization of richer sensory input (images from both wrist and static cameras plus proprioception) and predicting absolute joint positions rather than delta end-effector poses.

Trajectory Transfer completes the task on the orange puzzle in 9 of 10 trials. Furthermore, it demonstrates partial generalization to the green and blue puzzles, achieving insertion-stage success in 7 and 8 trials, respectively. However, because this method relies on tracking spatially adapted Cartesian trajectories, it inherently fails to accommodate puzzles with differing internal geometries during the completion stage. Additionally, it remains highly sensitive to visual alignment errors, which is shown in the Appendix~\ref{appendix:visualAugmentation}.

Despite varying degrees of success in the \emph{insertion} stage, none of the baseline methods were able to derive a policy that generalizes from the orange puzzle to other puzzles.

In contrast, our approach successfully accomplishes the task on the orange puzzle in all 10 trials and directly generalizes to the green and blue puzzles, as shown in \videotwo. This one-shot generalization stems from generating motions with respect to ECs in the contact space rather than mimicking exact demonstration actions. For example, our policy rotates the grasped part until it passes through a pin, regardless of the specific rotation angles performed (see Fig.~\ref{fig:PuzzleOpeningProcess}). As a result, a single demonstration is sufficient to reliably accomplish the task and generalize across geometric variations.

The efficient generalization of our approach highlights that ECs serve as a suitable abstraction, preserving the general task structure while discarding instance-specific details (e.g., absolute Cartesian trajectories), thereby enabling a policy learned on a single instance to directly generalize to novel geometries.

\begin{table}
	\centering
	\resizebox{1\linewidth}{!}{
		\begin{tabular}{c | c c | c c | c c}
			\hline
			\midrule
			Method & \multicolumn{2}{c|}{Orange} & \multicolumn{2}{c|}{Green} & \multicolumn{2}{c}{Blue} \\
			& Insertion & Completion & Insertion & Completion & Insertion & Completion \\
			\midrule
			BC~\cite{BC2} & 0/10 & 0/10 & 0/10 & 0/10 & 0/10 & 0/10 \\
			VINN~\cite{VINN} & 0/10 & 0/10 & 0/10 & 0/10 & 0/10 & 0/10 \\
			ACT~\cite{ACT} & 5/10 & 0/10 & 1/10 & 0/10 & 3/10 & 0/10 \\
			Diffusion Policy~\cite{DP} & 6/10 & 0/10 & 2/10 & 0/10 & 5/10 & 0/10 \\
			Trajectory Transfer~\cite{vitiello2023one} & 10/10 & 9/10 & 7/10 & 0/10 & 8/10 & 0/10 \\
			\textbf{Ours} & \textbf{10/10} & \textbf{10/10} & \textbf{10/10} & \textbf{10/10} & \textbf{10/10} & \textbf{10/10} \\
			\hline
		\end{tabular}
	}
	\caption{Success rates of different approaches on the puzzle insertion tasks. By extracting parameterized EC primitives from a single augmented demonstration and executing compliant motions within the contact space, our approach consistently outperforms all baseline methods and demonstrates efficient generalization to unseen puzzle geometries.}
	\label{table:Puzzle0Degree}
	\centering
\end{table}

\subsection{Door Locks with Keys}
\label{sec:doorlocks}

\begin{figure}
	\centering
	\includegraphics[width=1\linewidth]{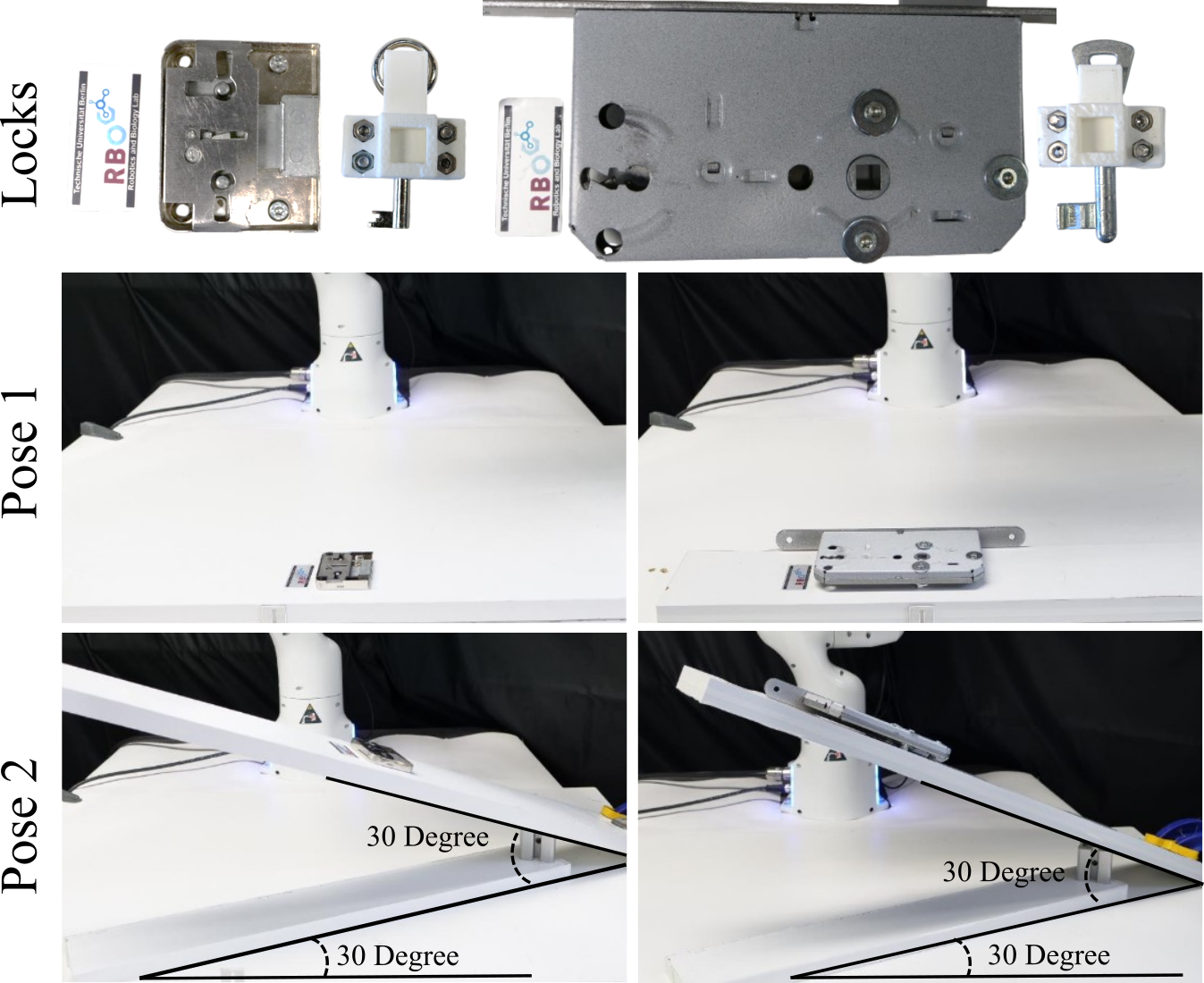}
	\caption{Experimental setup for the door-lock manipulation task. The policy extracted from Lock 1 is extended to Lock 2 via two targeted corrective demonstrations. We intentionally tilt the mounting surface to demonstrate that our approach does not rely on a tabletop assumption.}
	\label{fig:KeyExperimentSetup}
\end{figure}

We now turn to a more challenging task: door lock manipulation. Fig.~\ref{fig:KeyExperimentSetup} illustrates two locks used in the experiment. We first record a demonstration on Lock~1. Our approach segments this demonstration into three motion phases, augments them through exploratory actions to identify the EC types and contact-changing events for in-contact phases, and builds a policy composed of three EC primitives. The demonstration and augmentation process are illustrated in~\videofour$\,$.

We evaluate our policy on Lock 1 under two object poses with 10 trials each. A trial is defined as a success if the door lock is toggled. As seen in Table~\ref{table:KeyExperimentResults} and \videofive$\,$, our approach succeeded in all 20 trials. Notably, the robot was unable to insert the key by simply replaying the free-space and plane trajectories. Instead, the in-built exploration actions of the plane primitive, e.g., visual servoing and spiral search, are necessary for the successful insertion across all trials.

These results clearly demonstrate that tracking Cartesian positional trajectories is prone to failure for fine-grained contact-rich tasks.  By contrast, our approach utilizes EC primitives to exploit contact feedback, yielding robust execution.

We then evaluate this policy on Lock 2. As shown in Table~\ref{table:KeyExperimentResults}, direct application of the policy fails to insert the key. This failure is primarily caused by two factors. First, differences in object appearance lead to errors in the visual alignment. Second, the increased height of Lock 2 causes the robot to make premature contact with the surface, terminating the free-space primitive at an incorrect pose and propagating errors to the subsequent plane primitive. These instance-specific differences are not captured by our EC primitives. As a result, the policy fails to open Lock 2.

We address these failures using targeted human corrections (\videosix). After executing the plane primitive, the robot fails to observe the expected breaking-contact event that indicates successful insertion. This discrepancy is correctly recognized as an execution failure, triggering a request for human correction. We then provide a correction that guides the end-effector from the failed pose toward the keyhole. This correction is instantiated as an additional plane primitive with an updated demonstration trajectory and a new reference image for visual servoing. Evaluating the refined policy on Lock 2 yields improved success rates of 7/10 and 8/10 for Pose 1 and Pose 2, respectively.

The remaining failures arise from collisions between the key and the hole edge, which trigger an unexpected making-contact event that is then recognized as an execution failure. We then provide a second correction that introduces a wiggling motion, tilting and rotating the key around the hole, to resolve rotational misalignment. With this second correction, the policy succeeds in all 20 trials (see~\videosix).

This experiment underscores the critical importance of incorporating human corrections to manage out-of-distribution instance variations that are difficult to anticipate and model in the policy representation. Our approach detects such failures by monitoring contact transitions. Moreover, the modular and interpretable EC-based policy structure supports seamless integration of corrective demonstrations, enabling targeted adaptation to new task configurations without extensive retraining.

\begin{table}[!htbp]
	\centering
	\resizebox{1\linewidth}{!}{
		\begin{tblr}{
				colspec = {crrrr},
				cell{1}{2,4} = {c=2}{c}
			}
			\hline
			\hline
			Number of corrections &   Lock 1 &      &  Lock 2 &      \\
			\hline[0.8pt]
			& Pose 1 & Pose 2& Pose 1& Pose 2 \\
			\cline{2-5}
			No correction    &  10/10 & 10/10 & 0/10  & 0/10 \\
			1st correction   &  -     &   -   & 7/10 & 8/10 \\
			2nd correction  &   -    &   -   &  10/10 & 10/10 \\
			\hline
		\end{tblr}
	}
	\caption{Success rates for manipulating Lock 1 and Lock 2 under different numbers of human corrections.}
	\label{table:KeyExperimentResults}
	\centering
\end{table}

\begin{figure}
	\centering
	\includegraphics[width=1\linewidth]{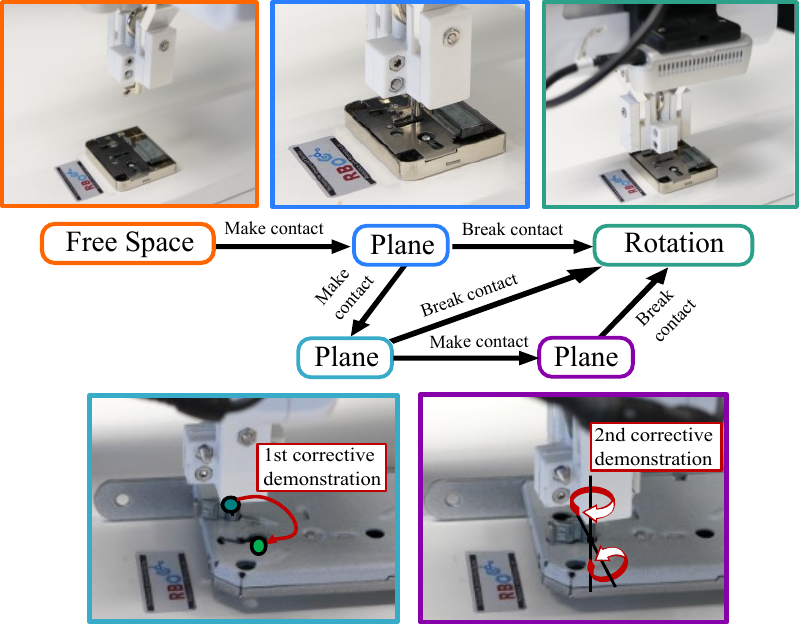}
	\caption{Policy refinement for door lock manipulation. The policy learned from Lock 1 is extended with two corrective demonstrations to handle unseen contact changes in Lock 2. The first correction guides the end-effector from an incorrect pose toward the keyhole, while the second correction introduces wiggling motions to resolve misalignment during insertion. Our EC-based policy representation enables seamless integration of these corrections as additional EC primitives without retraining the full policy.}
	\label{fig:PolicyForDoorLockWithKeys}
\end{figure}

\subsection{Coffee Machine}
\label{sec:coffeemachine}

\begin{figure}[htbp]
	\centering
	\includegraphics[width=\linewidth]{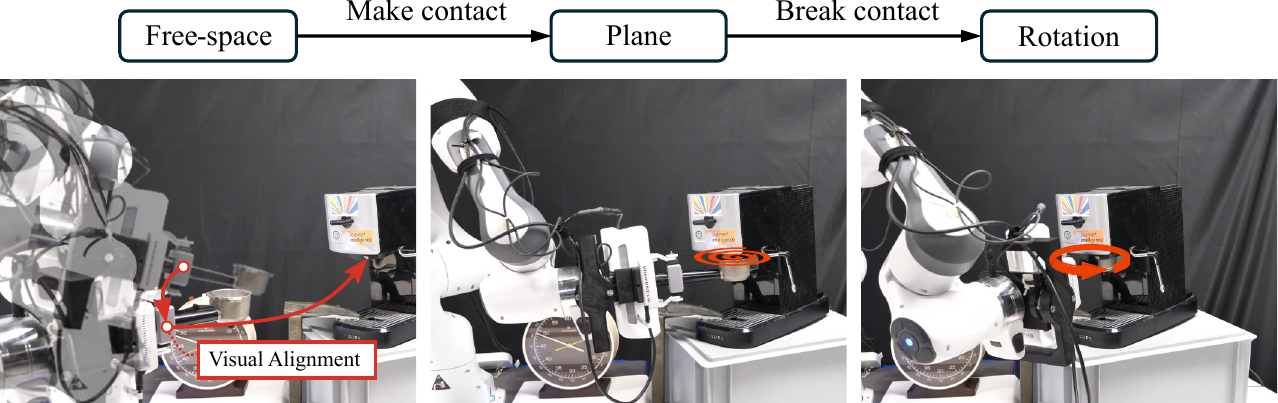}
	\caption{Execution of the coffee machine manipulation policy consisting of three EC primitives. First, the robot performs the visual alignment and executes a transferred trajectory. Upon making contact with the machine, the robot employs a plane primitive to align the portafilter with the insertion slot. Finally, upon detecting a contact-breaking event, it uses a rotation primitive to secure the portafilter.}
	\label{fig:coffeeMachinePolicy}
\end{figure}

We next evaluate our approach on a coffee machine manipulation task to demonstrate that online adaptation contributes to generalization to imprecise EC models. A policy composed of three EC primitives is constructed from a single augmented demonstration, as shown in Fig.~\ref{fig:coffeeMachinePolicy}. As shown in \videoExpCoffee$\,$, the policy reliably reproduces the demonstration across 10/10 trials, even under pose variations of the machine.

\begin{figure}[htbp]
	\centering
	\begin{subfigure}[b]{0.37\linewidth}
		\includegraphics[width=\linewidth]{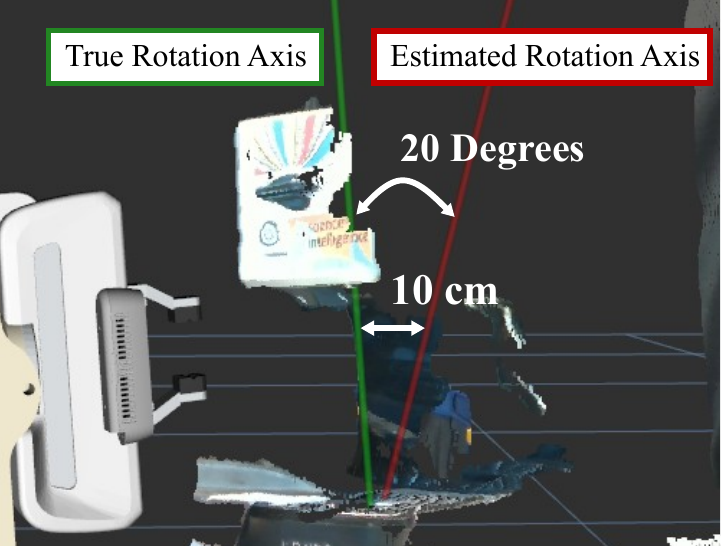}
	\end{subfigure}
	\begin{subfigure}[b]{0.61\linewidth}
		\includegraphics[width=\linewidth]{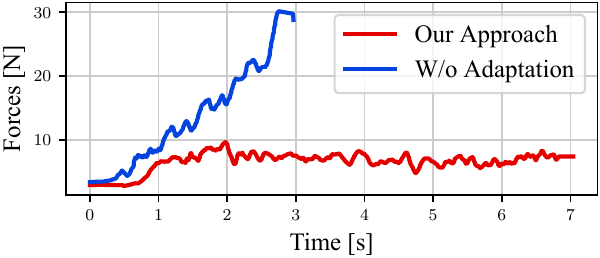}
	\end{subfigure}
	\caption{Evaluating policy performance under modeling errors. \textbf{Left:} We inject errors into the rotation EC model by applying a translational offset and an angular tilt to the rotation axis. \textbf{Right:} Our approach continuously updates the EC parameters based on the observed movement, successfully rotating the portafilter while maintaining low reaction forces. In contrast, executing the task without adaptation quickly exceeds the force limit, resulting in failure.}
	\label{fig:RobustToAxisErrors}
\end{figure}

Next, we introduce modeling errors into the EC model of the rotation primitive, shifting the rotation axis by \SI{20}{\degree} and \SI{10}{\cm} from the rotation center, as visualized in Fig.~\ref{fig:RobustToAxisErrors}. Our approach successfully corrects these errors by generating compliant motions and updating its EC model online based on observed movements (see \videoOnlineAdapt). Consequently, our method completes 9 out of 10 trials, whereas a baseline approach following a precomputed trajectory without online adaptation consistently fails, as the geometric errors rapidly drive the reaction forces beyond safe interaction limits (see Fig.~\ref{fig:RobustToAxisErrors}).

Overall, our approach reliably completes the task despite pose variations, reduced manipulability near joint limits, and imprecise EC models. It generalizes well because we do not attempt to learn complete EC models from a single demonstration. Instead, our primitives integrate sensory feedback to update models at run time. As we show, this online adaptation is essential for real-world physical interaction; even small modeling errors quickly generate large reaction forces that cause manipulation failures.

\subsection{Latches}
\label{sec:latches}

\begin{figure}[htbp]
	\centering
	\includegraphics[width=\linewidth]{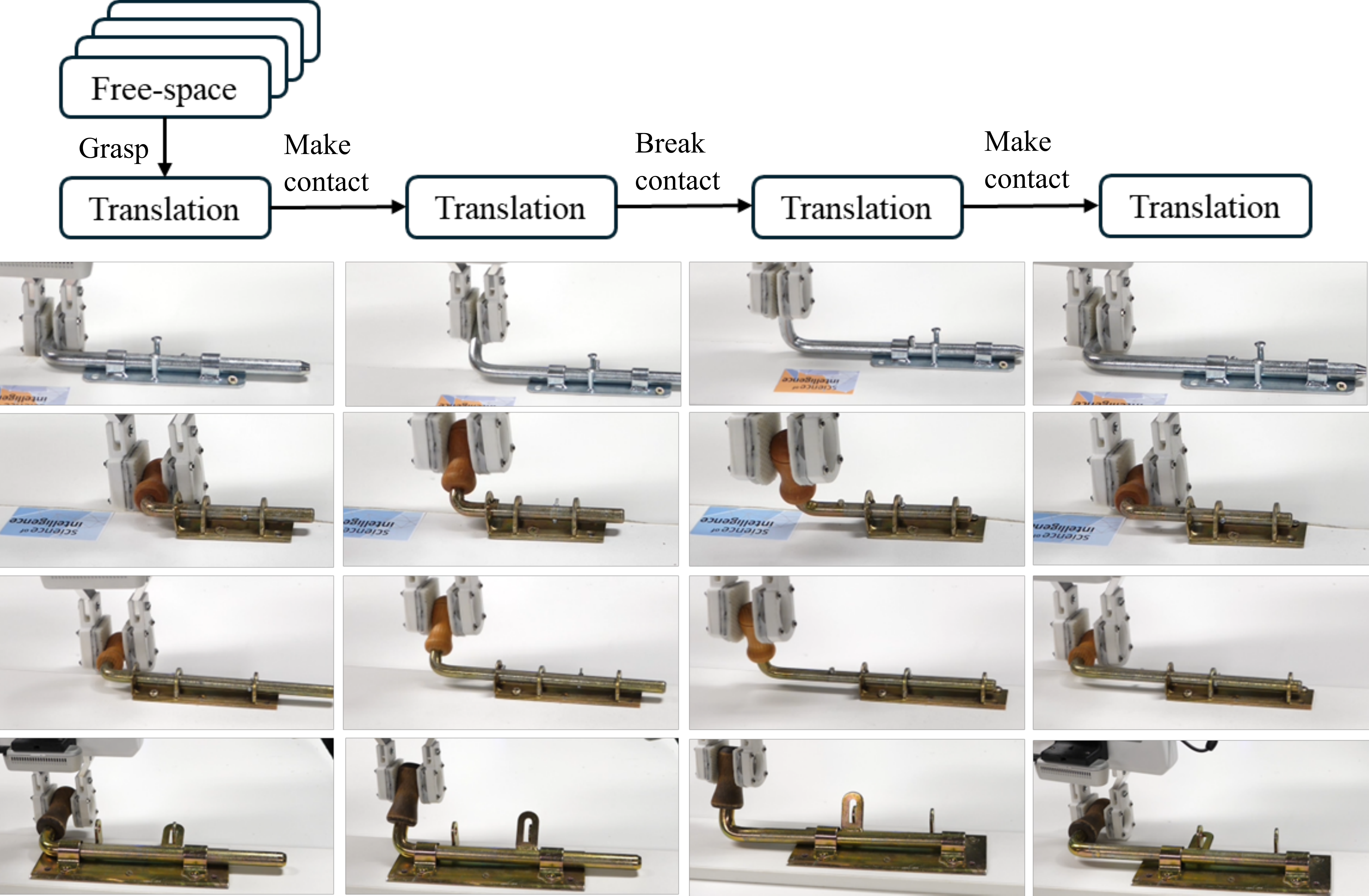}
	\caption{EC-based policy execution across different latches. The top row illustrates the policy structure composed of five EC primitives. The image sequences show successful execution on four different latches using a policy extracted from Latch~1 and adapted only with additional grasp demonstrations. By generating motions in the contact space, the policy generalizes to all latches varied in geometry.}
	\label{fig:latchPolicy}
\end{figure}

We then test our approach on complex mechanisms involving multiple DOFs. The first such task is latch manipulation, illustrated in Fig.~\ref{fig:latchPolicy}. Opening a latch involves a sequence of actions: grasping the knob, sliding it, lifting it until a pin passes a slot, sliding to the end position, and finally lowering it down.

We perform a kinesthetic demonstration on Latch~1 and segment it into five motion phases. After the augmentation, the robot builds a policy that consists of five EC primitives, as shown in Fig.~\ref{fig:latchPolicy}. Since all latches have distinct appearances, rather than relying on sophisticated vision algorithms for pose estimation, our approach efficiently extends the existing policy with one grasp demonstration as a correction per latch.

We run 10 trials per latch. For Latches~1–3, success is defined as the pin passing through the slot; for Latch~4, success is defined as the hole in the bar aligning with the protrusions on the plate.

The results are summarized in Table~\ref{table:LatchLocks} and can be seen in \videolatch$\,$. Although each latch requires its own grasping demonstration, the policy extracted from Latch~1 generalizes effectively, achieving 39 successful executions out of 40 trials across all four latches. The single failure was on Latch~4, where its reflective surface disrupted depth measurements, causing a grasp failure. This highlights the need for more robust keypoint extraction and matching algorithms, a limitation we discuss in further detail in Sec.~\ref{sec:limitations}.

Our method generalizes to new latches with different geometries by using ECs as the representation and generating motions in the contact space. Crucially, sliding and rotation motions terminate when contact-change events are detected, not when a specific distance or rotation angle is reached. Moreover, when lifting the knob, the translation primitive dynamically adjusts motion direction based on observed movement, allowing the robot to adaptively follow the structure of the constraints. This online adaptation enables generalization to unseen geometric structures and hard-to-model slippages at run time.

\begin{table}
	\centering
	\resizebox{0.9\linewidth}{!}{
		\begin{tabular}{c | c c c c}
			\hline
			\midrule
			Method  &Latch 1 & Latch 2 & Latch 3 & Latch 4\\

			\midrule
			Our approach & 10/10 & 10/10 & 10/10 & 9/10 \\
			\hline
		\end{tabular}
	}
	\caption{Success rates of grasping and opening latches. The only failure occurred in Latch~4 due to large visual alignment errors that resulted in a failed grasp.}
	\label{table:LatchLocks}
	\centering
\end{table}

\subsection{Drawers with Handles}
\label{sec:drawers}

\begin{figure}[htbp]
	\centering
	\includegraphics[width=\linewidth]{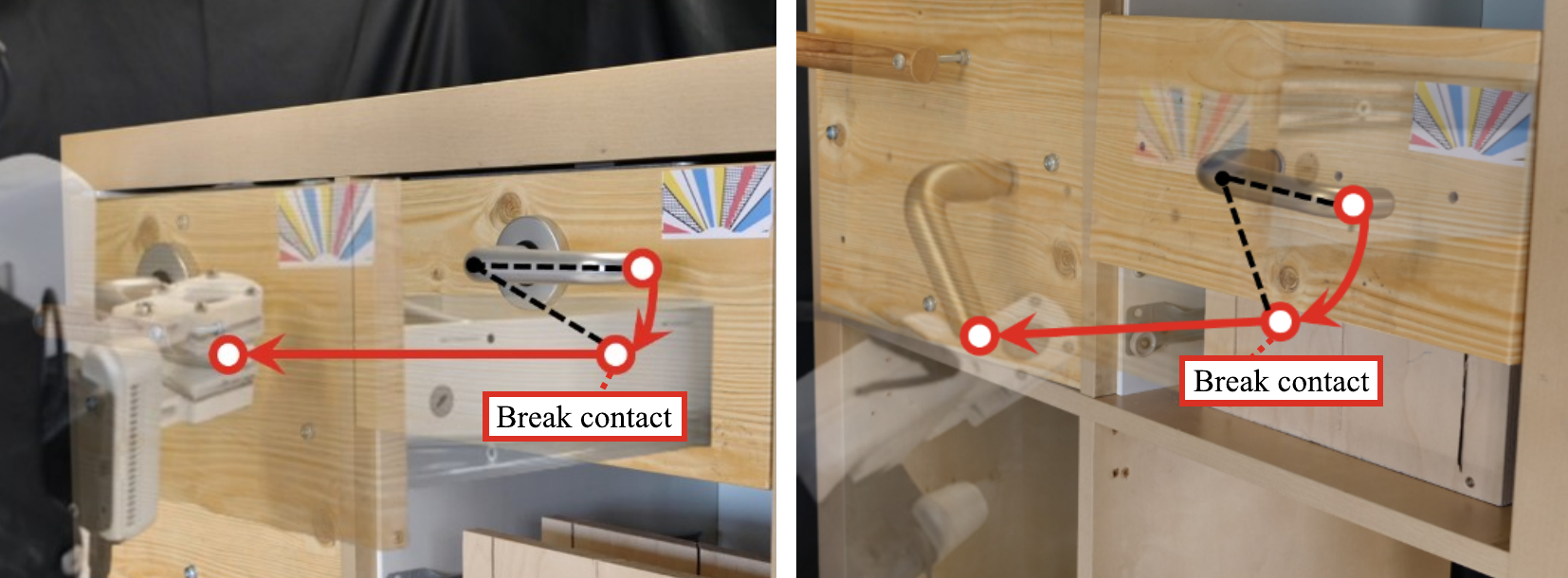}
	\caption{Execution of the policy of opening two drawers. By generating motions that follow the kinematic structure of the handle and drawer, a single policy reliably succeeds despite unseen variations in object pose and geometry.}
	\label{fig:drawerPolicy}
\end{figure}

Similar to the latch experiment, the drawer experiment again shows that ECs enable generalization across geometric variations. In this task, the robot must open two drawers with different handle rotation angles (see Fig.~\ref{fig:drawerPolicy}).

Our approach builds a policy from a kinesthetic demonstration on the drawer with the silver handle. The demonstration is segmented into three motion phases: grasping the handle, pushing it downward, and pulling the drawer. After augmentation, we obtain a policy consisting of one free-space primitive for grasping and two translation primitives connected by a breaking-contact event. The handle-rotation behavior is recognized as a translation primitive due to the small observed rotation motion. Nonetheless, the robot can still rotate the handle with a translation primitive by compliantly following the admissible motion direction estimated at run time.

We run 10 trials per drawer and consider a trial successful if the drawer is opened. Each drawer is affixed with a sticker to provide consistent SIFT features for the visual alignment. As shown in \videodrawer, our approach succeeds in all 20/20 trials across both drawers, despite differences in required rotation angles and variations in drawer poses.

These results further support our central claim: ECs separate generalizable contact information that captures the task structure while abstracting away task-irrelevant details, such as exact rotation angles. This separation simplifies the learning process and enables one-shot generalization to novel instances that vary in task-irrelevant properties.

\subsection{Functional Manipulation Benchmark}
\label{sec:FMB}

Finally, we apply our approach to two long-horizon assembly tasks~\cite{fmb2023, FB2023}, which involve multiple free-space and in-contact motions. As shown below, the complex behaviors required for these tasks can be effectively reproduced using a sequence of EC primitives, parameterized from a single augmented demonstration.

We start with the single-object insertion task from the Functional Manipulation Benchmark~\cite{fmb2023}. The robot needs to grasp the object, reposition it on a fixture for a regrasp, and finally insert it into the board. We provide a demonstration using the square-circle object. The demonstration is segmented into four free-space motion phases and two in-contact motion phases. After augmentation, we obtain a policy composed of six EC primitives, as shown in Fig.~\ref{fig:FMBPolicy}.

\begin{figure}[htbp]
	\centering
	\begin{minipage}{\linewidth}
		\centering
		\includegraphics[width=\linewidth]{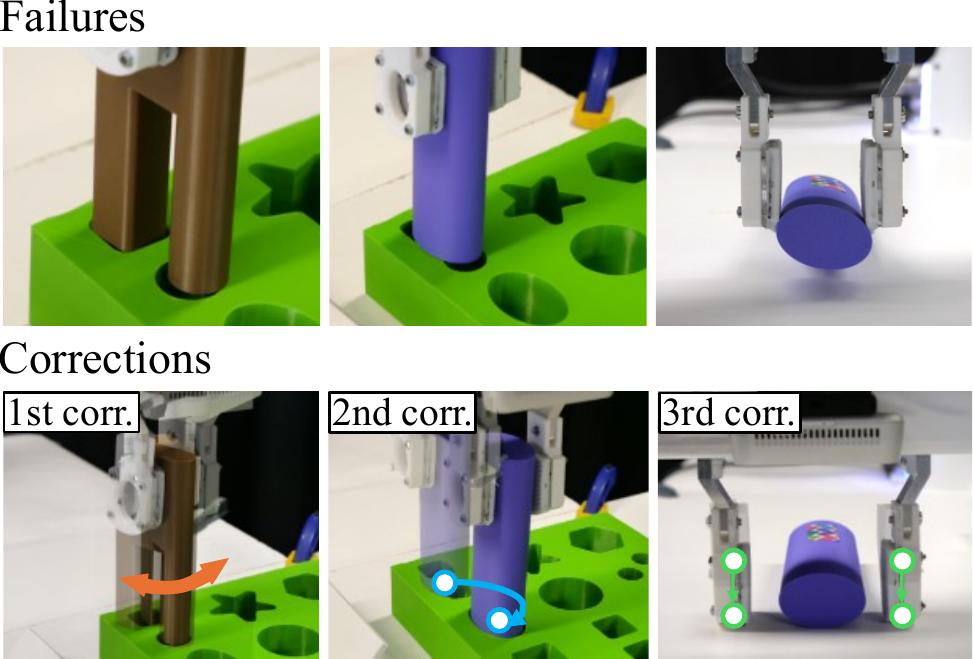}
		\captionof{figure}{Failures and corrections in the FMB assembly task. The top row shows three failure cases: object-board collision, wrong hole insertion, and lost grasp. The bottom row illustrates the corresponding human corrections: local wiggling to resolve misalignment, retargeting the correct insertion hole, and adjusting the grasp pose.}
		\label{fig:fmbCorrections}
		\vspace{1em} 
		\resizebox{1\linewidth}{!}{
			\begin{tabular}{c | c c c c}
				\hline
				\midrule
				Objects  & No corr. & 1st corr. & 2nd corr. & 3rd corr. \\
				\midrule
				Square-circle & 9/10 & 10/10 & - & - \\
				Oval & 0/10 & - & 3/10  & 9/10 \\
				\hline
			\end{tabular}
		}
		\captionof{table}{Success rates for inserting square-circle and oval objects into the board under different numbers of human corrections.}
		\label{table:FMBresults}
	\end{minipage}
\end{figure}

We test the policy in 10 trials using the same square-circle object, with varied poses and tilts of the plate. A trial is successful if the object is fully inserted into the board. The policy succeeds in 9/10 trials. The one failure is due to misalignment, causing the object to collide with the board and trigger a making-contact event. The robot detects this by monitoring contact-change discrepancies. We then provide a corrective demonstration involving a wiggling motion to resolve the misalignment, similar to the one used in the door lock task (Sec.~\ref{sec:doorlocks}). With this correction, the policy succeeds in all 10 trials, as shown in Table~\ref{table:FMBresults} and \videoFMBOne$\,$.

The policy with the square-circle object can be transferred to the oval object with two additional human corrections (see Fig.~\ref{fig:fmbCorrections}). When applying the policy to the oval object, the robot fails to detect a breaking-contact event after executing the plane primitive, because the oval object does not fit the square hole. A second corrective demonstration is then provided to guide the oval object toward the correct hole. With this correction, the policy succeeds in 3/10 trials.

\begin{figure}[htbp]
	\centering
	\includegraphics[width=\linewidth]{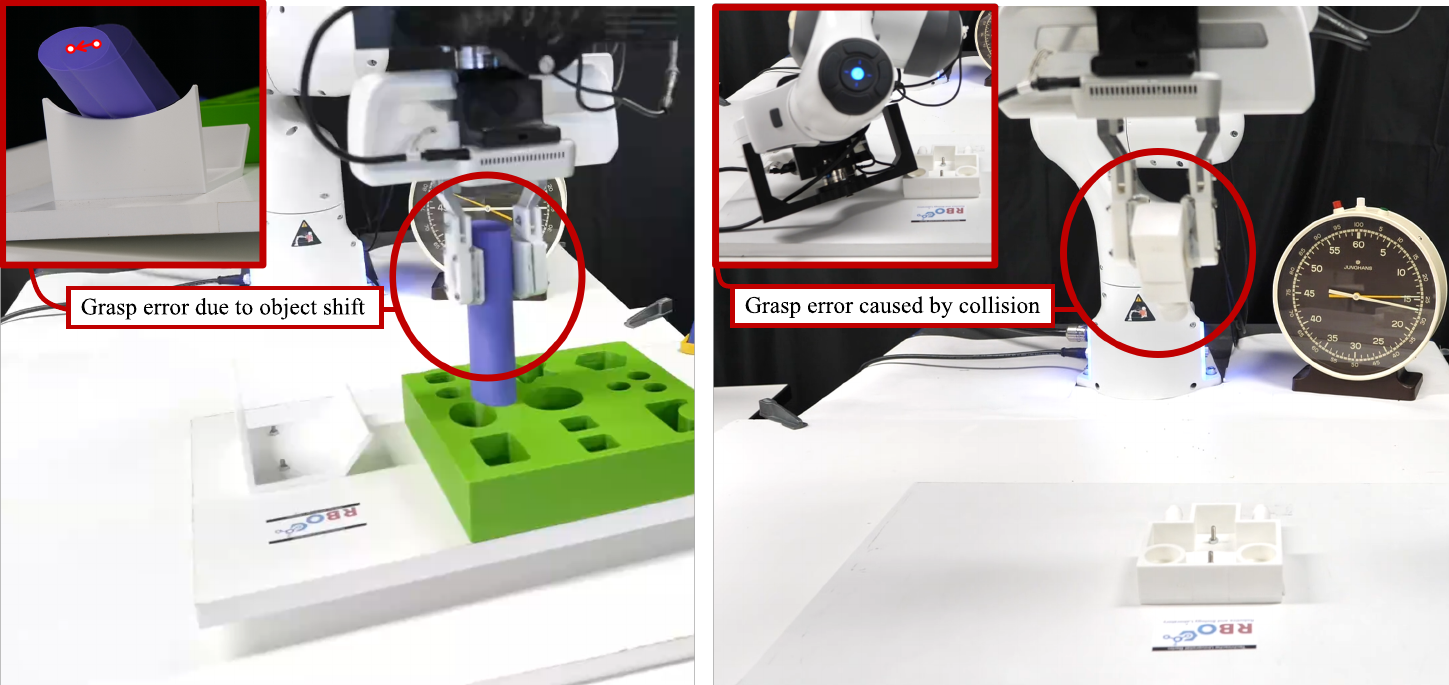}
	\caption{Two failure cases observed in the FMB and FB tasks due to incorrect grasp poses. \textbf{Left:} Tilting the plate causes the oval object to shift, leading to an inaccurate grasp. \textbf{Right:} During the execution of a free-space primitive, the robot collides with the insertion board, also resulting in errors in the grasp pose. These failures underscore the importance of integrating sensors for in-hand object pose estimation for precise assembly tasks.}
	\label{fig:failureCases}
\end{figure}

The remaining failures are due to lost grasps during execution. The grasp pose from the square-circle object does not transfer well to the oval object due to differences in their shape and thickness, leading to failures of a lost grasp. These failures are detected by monitoring the gripper state during the execution. We introduce a third correction that grasps the oval object. With both the second and third corrections applied, the policy succeeds in 9/10 trials, as can be seen in \videoFMBTwo$\,$. The one failure is caused by a large in-hand pose error after grasping, which is shown in Fig.~\ref{fig:failureCases} and discussed in Sec.~\ref{sec:limitations}.

\subsection{One-Leg Assembly Task}
\label{sec:FurnitureBench}

Our method is also applicable to the FurnitureBench benchmark~\cite{FB2023}. We evaluate it on the one-leg assembly task, where the robot grasps a leg, aligns it with a hole in the board, and screws it in.

\begin{figure*}
	\centering
	\begin{subfigure}{\textwidth}
		\includegraphics[width=\linewidth]{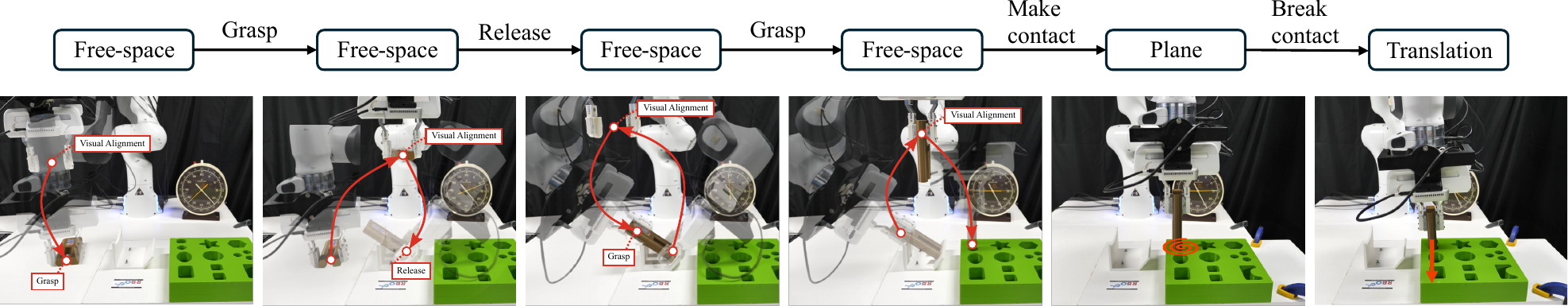}
		\caption{Execution of a single-object insertion policy consists of 6 EC primitives. \circled{1} The robot begins with a free-space primitive to grasp the object. \circled{2} It then executes a second free-space primitive: it replays part of the demonstration, performs visual realignment to the fixture, and follows the transferred trajectory to place the object on the fixture. \circled{3}–\circled{4} Two additional free-space primitives are executed to regrasp the object and approach the insertion board. \circled{5} Upon establishing contact with the board, the robot performs a plane primitive to align the object with the hole until a breaking-contact event is detected. \circled{6} Finally, it transitions to a translation primitive that compliantly inserts the object into the hole.}
		\label{fig:FMBPolicy}
		\vspace{2mm}
	\end{subfigure}
	\begin{subfigure}{\textwidth}
		\includegraphics[width=\linewidth]{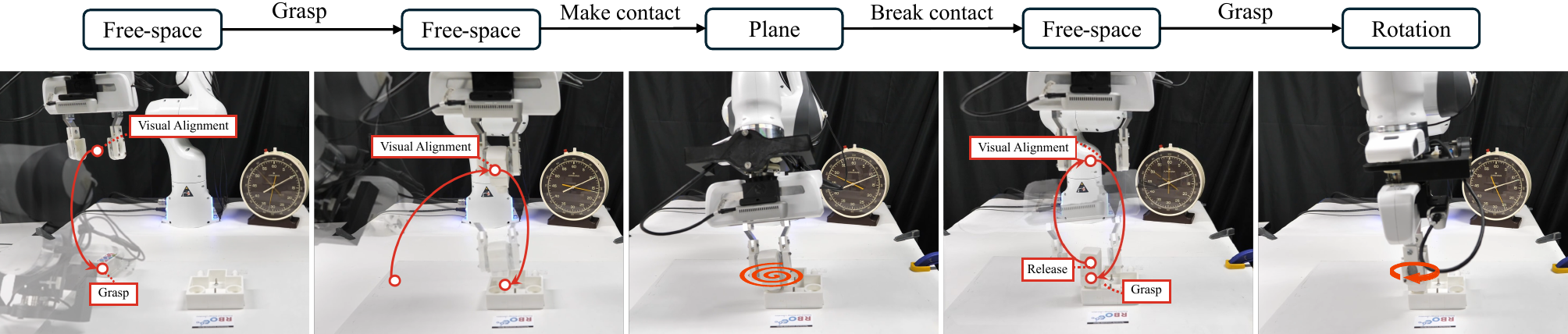}
		\caption{Execution of a one-leg assembly policy consists of 5 EC primitives. \circled{1} The robot starts with visually aligning to the leg and then executes an adapted free-space trajectory to grasp the leg. \circled{2} After grasping, it replays part of the demonstration trajectory, performs another visual alignment, and approaches the board to establish contact. \circled{3} The robot executes the plane primitive that aligns the leg with the hole on the board. \circled{4} Upon detecting a breaking-contact event, it transitions to a subsequent free-space primitive to regrasp the leg in a new pose. \circled{5} Finally, the robot completes the assembly by executing a rotation primitive to screw the leg into the board.}
		\label{fig:FBPolicy}
	\end{subfigure}

	\caption{Complex multi-stage contact-rich manipulation tasks can be decomposed into sequential motions. Our EC primitives are able to reproduce these EC-based motions from only one demonstration.}
	\label{fig:assembleGoalPoses}

\end{figure*}

Our approach segments a demonstration into five motion phases, followed by an augmentation process to recognize the underlying ECs and detect contact-changing events, and enhance visual features for visual alignment. The resulting policy consists of five EC primitives, as shown in Fig.~\ref{fig:FBPolicy}.

The demonstration, augmentation, and evaluation are shown in \videoFB$\,$. Despite substantial variations in object pose, the robot successfully completes 9/10 trials. The single failure is due to a collision between the robot and the insertion board, resulting in an incorrect grasp, as shown in Fig.~\ref{fig:failureCases}.

\subsection{Lessons Learned From Real-World Contact-Rich Manipulation}

Based on our extensive real-world experiments, we share several hands-on lessons that reveal the fundamental challenges in contact-rich manipulation. First, exact task trajectories, like the angle needed to turn a handle or open a latch, are nearly impossible to infer from visual observations alone. Second, when manipulating tightly constrained mechanisms, tiny perception or control errors cause massive reaction forces and jamming. These failure states are absent from demonstration data because skilled humans naturally avoid them. Third, even if the robot knows the correct motion and force profile, complex physical interactions, like continuous slippage between the gripper and the object, can still cause trajectory tracking to fail. Finally, in long-horizon tasks like the Functional Manipulation Benchmark and FurnitureBench, these errors compound, and a single failure would already fail the entire sequence. This explains why policies trained end-to-end on hundreds of demonstrations still struggle on these benchmarks~\cite{fmb2023, FB2023}. Together, these practical realities make it hard to learn a policy that reliably operates on the original demonstration object, let alone generalizes to new task settings.

In contrast, our results show that these complex behaviors are fundamentally \emph{simple}. Following our four-step pipeline, we first leverage a single demonstration and subsequent self-guided augmentation to extract a generalized policy structure based on ECs. We then rely on online adaptation via compliant control to recover instance-specific details and human corrections to handle out-of-distribution failures. Taken together, the aforementioned challenges are effectively resolved by building an EC-based policy through this pipeline. Without ECs, learning these tasks is unnecessarily difficult.

These findings naturally raise a central question: What is the more promising path to generalization, scaling up data or identifying the right inductive biases? We discuss this question in greater depth in Sec.~\ref{sec:dietersection}.

\section{Limitations and Future Work}
\label{sec:limitations}

Although the previous experiments demonstrate that our LfD approach generalizes effectively from a single demonstration per task, several key limitations remain for future work to address.

\subsection{Application Domains}

The strong one-shot generalization of our approach is mainly attributed to the EC-based representation. While we have successfully applied this representation to a range of contact-rich manipulation tasks, it remains domain-specific. First, our primitives focus on geometric constraints rather than exact Cartesian trajectories. This works for many rigid-object manipulation tasks, but would fail when the specific motion shape is the goal, such as writing or drawing. Second, our system requires discrete changes in the contact space as transitions. While this makes the robot less dependent on exact positions, it prevents the system from handling tasks with continuous contact or changing materials, such as peeling, grinding, or manipulating soft objects.

To broaden the applicability, we should expand our approach with more primitives. For example, we should incorporate trajectory-centric primitives, which extract task-relevant motion characteristics from demonstrations~\cite{DMP, trajPrimitive1, trajPrimitive2, trajPrimitive3, trajPrimitive5, zhou2016learningDMPForce3, trajPrimitive7, zhou2019learning}. In addition, it is desired to include primitives based on end-to-end methods as fallback mechanisms when the predefined primitives are not suited for a given task~\cite{luo2024multistage, xue2025reactive}. A selection mechanism is needed to coordinate between different types of primitives.

\subsection{Technical Limitations}

In addition to application domains, our approach has several limitations in the technical implementation.

First, we assume the demonstration can be segmented into a sequence of ECs. However, our segmentation can fail if the demonstration contains exploratory motions that do not follow any particular EC. We need an interactive segmentation algorithm where the robot refines segment boundaries using the concept of augmentation and human corrections. This will extend the generalization across different demonstrators.

Second, our free-space primitive relies on SIFT features, which fail on textureless objects or significant appearance variations. We plan to improve this by integrating learnable feature extraction methods that can be fine-tuned using our vision-based augmentation pipeline~\cite{learnableKey1, learnableKey2, learnableKey3}. Additionally, we should use keypoints that generalize better across different objects. Possible solutions include using shape-based keypoints for similar geometries~\cite{keyshape1, keyshape2, wang2025skil, kamil2025} or Vision-Language Models that match keypoints based on functional affordances~\cite{keypointLLM1, keypointLLM2, di2024dinobot, seker2025grounded}.

Third, failure cases observed in both benchmarks indicate that our current system is not robust to collisions and disturbances that perturb the grasp pose (see Fig.~\ref{fig:failureCases}). Future work should consider integrating a collision-avoidance module into existing primitives~\cite{collision1, collision2, collision3, collision4} and extending the system with sensors for in-hand object pose estimation to compensate for errors in grasp poses~\cite{iHP1, iHP2, iHP3, chen2025robust}.

Fourth, our method requires parameter tuning for specific objects, as shown in Table~\ref{tab:ParametersForLatchDrawerAndDoorLocks}. For example, we need to increase the wrench limit because latches involve higher friction than other objects. Similarly, drawer handles have internal springs and thus require higher pulling forces. For door locks, the threshold for detecting breaking-contact events must be reduced to accommodate the tighter mechanical tolerances. While some of these parameters could be directly inferred from demonstrations, a more general solution is to identify these parameters through, for example, active exploration~\cite{johannsmeier2019framework}, Reinforcement Learning~\cite{HEP_RL1, vuong2021learning}, or human corrections~\cite{HEP_HC1, HEP_HC2}. This would allow the policy to generalize more easily to new objects.

\begin{table}
	\centering
	\renewcommand{\arraystretch}{1.4} 
	\caption{Changed parameters for manipulating various objects}
	\label{tab:ParametersForLatchDrawerAndDoorLocks}
	\resizebox{\linewidth}{!}{%
		\begin{tabular}{lcccc}
			\hline
			\textbf{Param.} & \multicolumn{4}{c}{\textbf{Value}} \\
			\cline{2-5}
			& \textbf{Door Locks} & \textbf{Coffee Machine} & \textbf{Latches} & \textbf{Drawers} \\
			\hline
			$F_\mathrm{m}$ & [5, 5, 5, 2, 2, 2] & [5, 5, 5, 2, 2, 2] & [10, 10, 10, 2, 2, 2] & [25, 25, 25, 2, 2, 2] \\
			\hline
			$\mathrm{MCE}^\mathrm{p}$ & \SI{5}{\N} & \SI{8}{\N} & \SI{5}{\N} & \SI{5}{\N}\\
			\hline
			$\mathrm{BCE}$ & \SI{0.7}{\cm} & \SI{0.8}{\cm} & \SI{1.5}{\cm} & \SI{2}{\cm} \\
			\hline
		\end{tabular}
	}
\end{table}

\section{Better for Generalization: More Data or the Right Inductive Biases?}
\label{sec:dietersection}

We have presented extensive experimental evidence that using the right inductive bias---environmental constraints---produces outstanding generalization in contact-rich manipulation tasks. But there is also strong empirical support for data scaling as a way to achieve generalization. This evidence is most compelling in natural language processing~\cite{achiam2023gpt, touvron2023llama, bai2023qwen} and computer vision~\cite{radford2021learning, he2022masked, kirillov2023segment, oquab2024dinov2}. In these domains, task-specific inductive biases, such as the ones we use here, are \textit{avoided} in favor of training directly from massive multi-task datasets. The appeal of this strategy lies in its ability to extract broadly applicable rules across diverse tasks. It is frequently argued that data scaling will eventually also solve generalization in robotics~\cite{Data2025}. This seems to contradict our results that demonstrate superior generalization \textit{with} inductive biases.

In this section, we want to resolve this apparent contradiction by explaining that the problems encountered in robotics are fundamentally different from those in computer vision and natural language processing. We therefore have little justification for hoping that the game-changing results from other domains will simply transfer to robotics. Worse, robotics might represent one of the most challenging problems for these learning methods.

There are at least two reasons that make learning contact-rich manipulation difficult. First, the function we might attempt to learn is basically chaotic, making it very challenging to learn or to model. When considering contacts, small changes to initial conditions lead to drastically different outcomes, including drastically different sensory traces. When multiple, interacting contact points are involved, things get worse. There is little hope to learn such a function---this function might not actually ``exist'' in a stable form as minor changes to environmental conditions can lead to drastic changes. And even if we could learn this function, the generalization of this function should be rather limited, again, as a result of small changes having \textit{different} drastic consequences not seen during training. In contrast, by building the exploitation of environmental constraints into our approach, all motions seek a contact regime that is easy to maintain and detect. Environmental constraints as an inductive bias eliminate this problem, leading to better generalization.

The second reason is that meaningful training data, direly required for the scaling-data approach, is challenging to come by. This data is most likely to be generated by humans. But humans have such extensive manipulation experience, they are so skilled at such tasks, that the training data we can obtain by observing them is heavily shifted towards small regions of the action space. No matter how many data points (demonstrations) we obtain, it seems highly unlikely that we can cover in any meaningful way those parts of the state space that a robot might encounter during policy execution under different environmental conditions. Again, environmental constraints as an inductive bias resolve this issue by abstracting away task-irrelevant details (e.g., exact motion and force profiles). With this abstraction, we represent what used to be a large regions of the state space as a clearly identifiable lower-dimensional manifold. This is valid because the entire original region represents an equivalence class \textit{with respect to the environmental constraint}. This abstraction makes it easy to gather sufficient training data to cover the task-relevant space (one human demonstration augmented by the robot is sufficient in many cases).

While environmental constraints solve these two challenges effectively, one might still wonder whether such inductive biases could be automatically extracted from data. We believe this is rather unlikely, as shown in several recent works that successful task completion of end-to-end manipulation policies often reflects interpolation within training data rather than meaningful generalization~\cite{he2025demystifying, fei2025libero}. The same is true in natural language processing and computer vision, where effective inductive biases---like transformers and CNNs---did not arise from scaling but were intentionally designed to encode symmetry and invariance~\cite{Data2025}.

The arguments in favor of inductive biases made above in the context of this paper and supported by the extensive experimental evidence presented in Sec.~\ref{sec:experiments} seem to transfer to other problems in robotics. In robotics, we have to deal with the ``real world,'' i.e., with worlds for which no simple models exist, worlds that are chaotic or near chaotic, such as the world of complex and uncertain contact situations. We therefore believe that it is imperative in robotics to continue to rely on strong and task-specific inductive biases. Learning on top of such biases continues to be a fruitful and desirable path towards more competent robotic systems.

\section{Conclusion}

We presented a novel LfD approach that achieves one-shot generalization for various contact-rich manipulation tasks. Our approach achieves such generalization by extracting the task-general structure from a single demonstration, augmenting this representation through self-guided exploration, refining it via targeted human corrections, and instantiating instance-specific details online using sensory feedback. Extensive experiments across seven real-world, contact-rich manipulation tasks demonstrate that policies constructed through this four-step pipeline are substantially more general than those derived from trajectory-centric or purely data-driven methods. Furthermore, based on extensive hands-on experimentation, we provide key insights into why contact-rich manipulation remains difficult to learn and show that environmental constraints serve as a highly effective inductive bias to overcome these challenges. We believe that insights into achieving generalization can inspire future research on designing learning algorithms that exploit environmental constraints in different ways, enabling more principled and general LfD for contact-rich manipulation.

\begingroup
\footnotesize
\bibliographystyle{IEEEtran}
\bibliography{references}
\endgroup

\section*{Appendices}
\label{appendix}

\subsection{Generalization Across Lighting Conditions}
\label{appendix:visualAugmentation}

\begin{figure}
	\centering
	\includegraphics[width=1\linewidth]{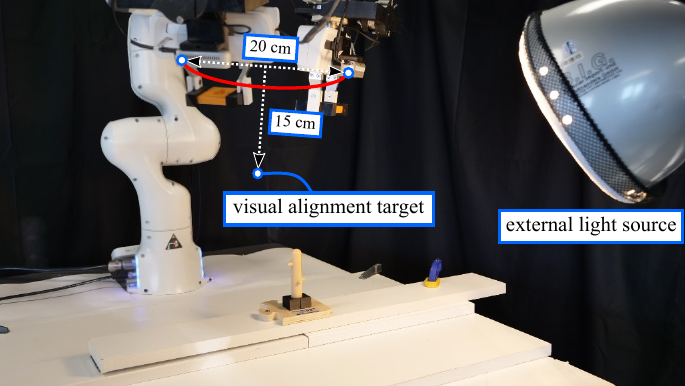}
	\caption{Experimental setup for evaluating the impact of vision-based augmentation under lighting changes. The robot performs visual alignment to return to a target pose, starting from 10 initial poses arranged along a circular trajectory.}
	\label{fig:lightingChangeSetup}
\end{figure}

This additional experiment shows how vision-based augmentation and EC exploitation support policy execution under increased visual alignment errors. To do so, we introduce an external light source to change lighting conditions (see Fig.~\ref{fig:lightingChangeSetup}).

\begin{figure}
	\centering
	\begin{subfigure}[b]{0.49\linewidth}
		\includegraphics[width=\linewidth]{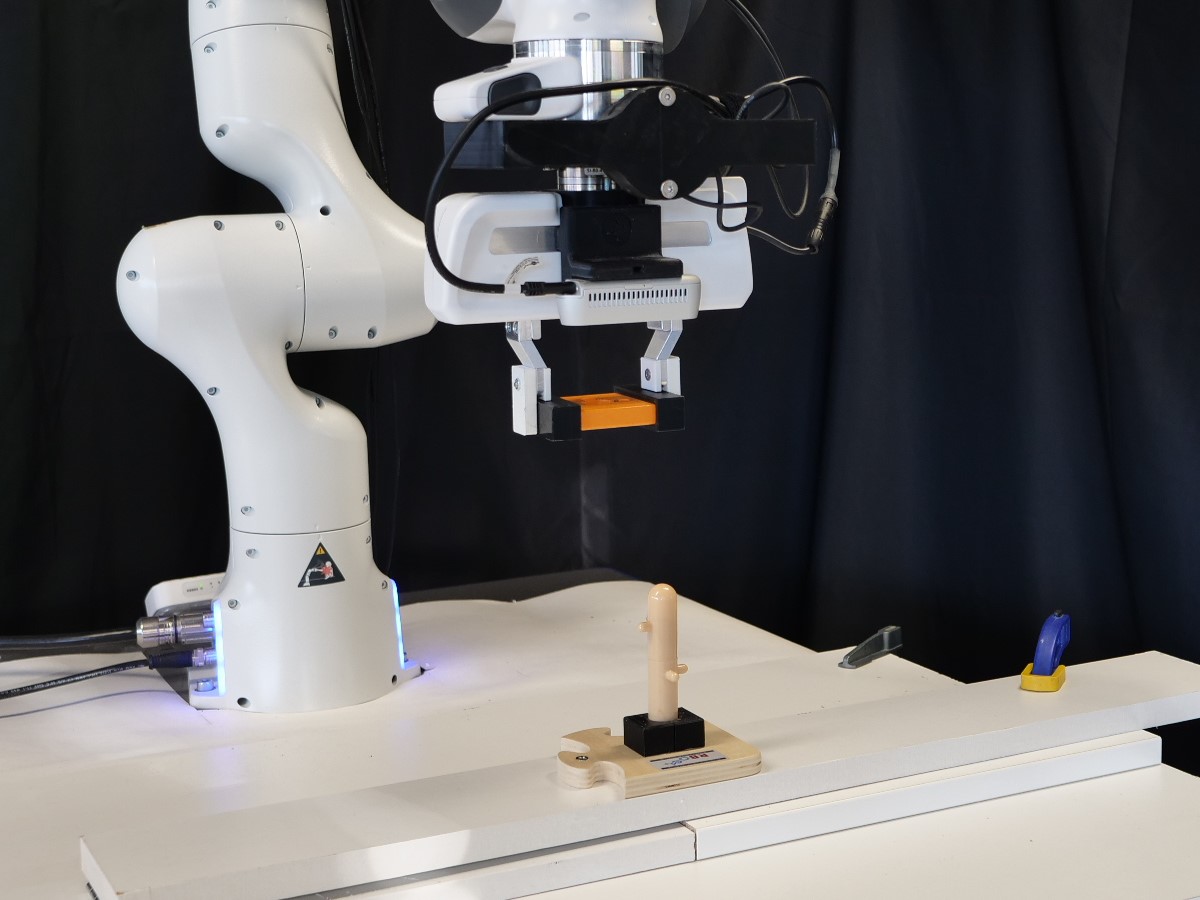}
	\end{subfigure}
	\begin{subfigure}[b]{0.49\linewidth}
		\includegraphics[width=\linewidth]{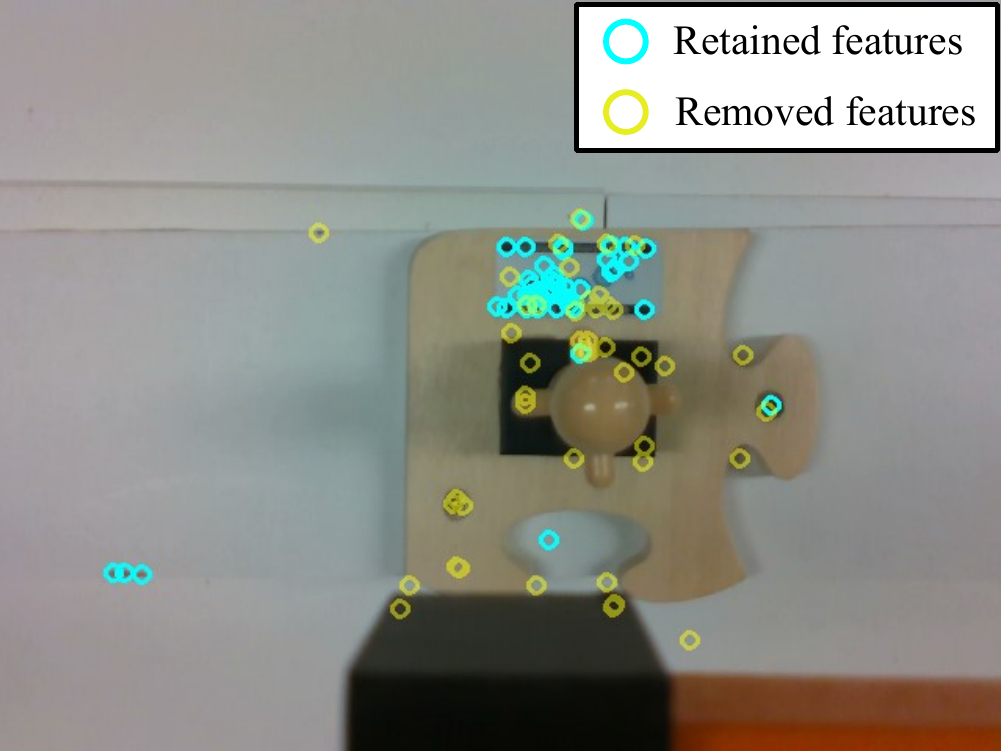}
	\end{subfigure}
	\caption{Effect of vision-based augmentation on the insertion puzzle task. \textbf{Left:} The robot at the pose where vision-based augmentation is applied. \textbf{Right:} Features on textureless regions or object boundaries, which are prone to mismatches or viewpoint-inconsistent, are removed by our vision-based augmentation strategy, leading to a better visual alignment performance.}
	\label{fig:vsAugmentationPuzzle}
\end{figure}

\begin{figure}
	\centering
	\includegraphics[width=1\linewidth]{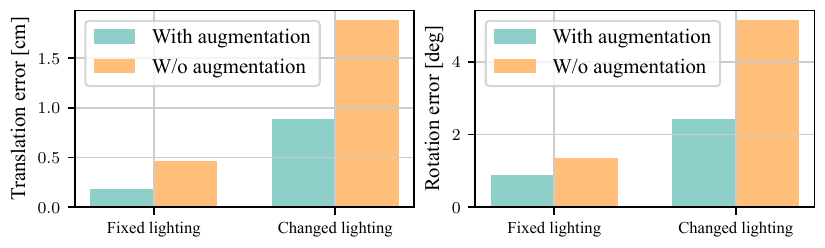}
	\caption{Comparison of average translational (left) and rotational (right) alignment errors in changed lighting conditions. The relatively small increase in error when lighting changes demonstrates the effectiveness of our vision-based augmentation strategy in enhancing the robustness of visual alignment to unseen lighting conditions.}
	\label{fig:vsalignmentErrors}
\end{figure}

We begin by moving the robot to a target pose, capturing a reference image, and recording the end-effector pose as ground truth. We then apply vision-based augmentation at this pose to improve image feature quality for the visual alignment (see Fig.~\ref{fig:vsAugmentationPuzzle} and \videoVSAug). Next, we turn on an external light to change the lighting conditions and test whether the robot can return to the target pose. We define 10 initial poses in a circular path \SI{15}{\cm} above the target and attempt alignment from each. We then compute translational and rotational errors of the end-effector pose relative to the ground truth. As shown in Fig.~\ref{fig:vsalignmentErrors}, alignment errors increase significantly when vision-based augmentation is omitted. These results confirm that the vision-based augmentation strategy enhances the visual alignment performance.

We next examine how increased alignment errors affect policy execution. We evaluate two methods that rely on visual alignment: trajectory transfer and our approach. Both methods use the same policies from the previous experiment, but are now tested under a modified task setup with changes in object pose and lighting conditions.

As shown in Table~\ref{table:ChangeLights}, Trajectory Transfer succeeds in only 3 of 10 trials without vision-based augmentation due to its sensitivity to visual alignment errors. We observed that even small orientation errors in the visual alignment can significantly disrupt the trajectory transfer. With augmentation, success improves to 60\% (6/10), consistent with the reduced alignment errors in Fig.~\ref{fig:vsalignmentErrors}.

Unlike Trajectory Transfer, our method with vision-based augmentation successfully completes all 10 trials (see \videothree). This high success rate is attributed to EC exploitations that effectively mitigate visual alignment errors. Specifically, as illustrated in Fig.~\ref{fig:sequenceECEs}, after executing a free-space primitive, the robot purposefully moves toward the puzzle board to establish contact, thereby reducing positional errors. Then, the plane primitive further compensates for remaining errors by performing visual servoing followed by a spiral search along the surface, ensuring robust execution.

\begin{table}
	\centering
	\resizebox{0.9\linewidth}{!}{
		\begin{tabular}{c | c c}
			\hline
			\midrule
			Method  &W/o augmentation & With augmentation\\
			\midrule
			Trajectory Transfer & 3/10 & 6/10  \\
			\textbf{Ours} & \textbf{9/10} & \textbf{10/10}  \\
			\hline
		\end{tabular}
	}
	\caption{Evaluation of trajectory transfer and our approach in a changed lighting condition, where the visual alignment errors increase (see Fig.~\ref{fig:vsalignmentErrors}). Our approach benefits from EC primitives that correct visual alignment errors through contact exploitation, achieving higher success rates. It also shows that vision-based augmentation increases the success rates of both approaches by increasing the performance of the visual alignment.
	}
	\label{table:ChangeLights}
	\centering
\end{table}

These results further underscore the critical role of incorporating ECs into the learning system. By leveraging ECs, the system can augment demonstrations with additional information and exploit contact feedback to reduce uncertainty, enabling robust execution under varying environmental conditions.

\end{document}